\documentclass[a4paper,fleqn]{cas-dc}

\usepackage[authoryear,longnamesfirst]{natbib}
\usepackage{todonotes}
\usepackage{enumitem}
\usepackage{longtable}

\newcommand{\mys}{\texttt{s-sum}}
\newcommand{\myt}{\texttt{t-sum}}
\newcommand{\ind}{\texttt{ind}}

\usepackage{algorithm}
\usepackage{amsmath}
\usepackage{algpseudocode}
\usepackage{amssymb}
\usepackage{nomencl}
\makenomenclature
\usepackage{float}
\usepackage{placeins}

\usepackage{graphicx}
\usepackage{booktabs}
\usepackage{multirow}
\usepackage{subcaption}
\usepackage[justification=centering]{caption}
\usepackage{url}
\usepackage{tabularx} %

\usepackage{amsthm}
\newtheorem{theorem}{Theorem}

\begin{document}
\let\WriteBookmarks\relax
\def\floatpagepagefraction{1}
\def\textpagefraction{.001}

\shorttitle{Probabilistic Forecasting and
Scenario Generation}

\shortauthors{Hanyu Zhang~et al.}

\title [mode = title]{Weather-Informed Probabilistic Forecasting and Scenario Generation in Power Systems}

\author[1,2]{Hanyu Zhang}[
                        bioid=1,
                        orcid=0000-0002-5903-0883]

\cormark[1]
\ead{hanyzhang@gatech.edu}
\affiliation[1]{organization={H. Milton Stewart School of Industrial and Systems Engineering, Georgia Institute of Technology},
    city={Atlanta},
    postcode={30332}, 
    state={GA},
    country={USA}}
    
\affiliation[2]{organization={AI Institute for Advances in Optimization},
    city={Atlanta},
    postcode={30332}, 
    state={GA},
    country={USA}}

\author[1,2]{Reza Zandehshahvar}[orcid = 0000-0002-6249-905X]
\ead{reza@isye.gatech.edu}

\author[1,2]{Mathieu Tanneau}[
orcid = 0000-0003-1554-8303
]
\ead{mathieu.tanneau@isye.gatech.edu}

\author[1,2]{Pascal {Van Hentenryck}}[
orcid = 0000-0001-7085-9994
   ]
\cormark[2]
\ead{pascal.vanhentenryck@isye.gatech.edu}

\cortext[cor1]{Corresponding author}
\cortext[cor2]{Principal corresponding author}

\nonumnote{This research was partly supported by NSF award 2112533 and ARPA-E PERFORM award AR0001136.}

\begin{abstract}
The integration of renewable energy sources (RES) into power grids presents significant challenges due to their intrinsic stochasticity and uncertainty, necessitating the development of new techniques for reliable and efficient forecasting. This paper proposes a method combining probabilistic forecasting and Gaussian copula for day-ahead prediction and scenario generation of load, wind, and solar power in high-dimensional contexts. By incorporating weather covariates and restoring spatio-temporal correlations, the proposed method enhances the reliability of probabilistic forecasts in RES. Extensive numerical experiments compare the effectiveness of different time series models, with performance evaluated using comprehensive metrics on a real-world and high-dimensional dataset from Midcontinent Independent System Operator (MISO). The results highlight the importance of weather information and demonstrate the efficacy of the Gaussian copula in generating realistic scenarios, with the proposed weather-informed Temporal Fusion Transformer (WI-TFT) model showing superior performance.

\end{abstract}


\maketitle

\section{Introduction}
\label{sec:introduction}
The adoption of Renewable Energy Sources (RES), especially wind and solar generation, is at the forefront of modern power systems' transition towards sustainable and carbon-free electricity generation.
This transition also promises to improve the affordability and accessibility of energy systems worldwide \citep{doe_renewable_systems_integration,doe_distributed_wind}. 
However, integrating high levels of RES into existing power grids introduces significant challenges, particularly in forecasting.
Indeed, RES, whose output depends on weather conditions, are intrinsically intermittent and stochastic.
This operational uncertainty thus complicates grid stability and optimization efforts \citep{doe_renewable_systems_integration} which, in turn, motivates the need for advanced forecasting tools.

In that context, academics and practitioners have increasingly moved from deterministic to \emph{probabilistic} forecasting, which better captures the uncertainty in load and RES output.
In addition, probabilistic forecasts provide the ability to sample multiple scenarios, which is essential for risk quantification \citep{Stover2023_RiskMetrics} and uncertainty-aware optimization \citep{Knueven2023_StochasticLAC}.
Despite their popularity, probabilistic forecasts present substantial scalability challenges.
Indeed, they must capture \emph{spatial} (between system elements located at different locations), and \emph{temporal} (between different time periods) correlations.
Typical forecasting tasks comprise hundreds of renewable generators and dozens of time steps (e.g. 48 hours), with output dimensions in tens of thousands, and correlation matrices with hundreds of millions of coefficients.

To address these challenges, this paper focuses on the development of high-dimensional probabilistic forecasting methods for load and RES generation.
By utilizing weather information as covariates and restoring the spatio-temporal correlations among the prediction variables, this paper sets the foundation for reliable scenario generation using high-dimensional multivariate time series forecasting in RES.
\subsection{Related Work}
\vspace{0.1cm}
\subsubsection{Time Series Forecasting}
Historically, the domain of time series prediction was dominated by statistical methods, with Autoregressive Integrated Moving Average (ARIMA) \citep{arima} model being a prominent example. However, these methods often fall short in capturing the non-linear and non-stationary nature of the time series and might not be reliable for forecasting RES. The advent of Recurrent Neural Networks (RNNs) marked a significant shift, demonstrating superior capabilities in capturing complex temporal dependencies inherent in time series data \citep{sutskever2014sequence,smyl2020hybrid}. Notably, the sequence-to-sequence (S2S) RNN architecture \citep{sutskever2014sequence}, initially developed for neural machine translation, has inspired the development of innovative time series forecasting models such as multi-quantile RNN \citep{wen2017multi}, DeepAR model \citep{salinas2020deepar}, and Temporal Fusion Transformer (TFT) \citep{LIM20211748}. By leveraging cross-series and cross-temporal learning, these methods have gained promising performance across diverse fields, which motivates their utilization for RES forecasting \citep{8657666,8715887,srivastava2020predictive}.  For a comprehensive survey of the recent advancements in RNN-based and general deep learning architectures in time series forecasting, readers are referred to \cite{hewamalage2021recurrent} and \cite{lim2021time}. 

Despite their significant advancements, RNN-based models and their derivatives encounter substantial challenges in accurately forecasting the dynamics of modern power systems. These challenges stem largely from the intermittency, uncertainty, and stochasticity associated with the integration of RES and distributed energy resources into existing power networks \citep{8113587}. Such complexities highlights the urgent need to pivot from the deterministic forecasting approaches to uncertainty quantification of forecast errors \citep{pinson2007non,bremnes2004probabilistic}. Many studies have adopted parametric approaches to model the distribution of forecast errors, utilizing distributions such as the beta distribution \citep{beta_wind_power} and the logit-normal distribution \citep{lognormal_wind_power}. These approaches, while valuable, often presume a specific functional form for the forecast errors, which may not always align with the observed data. In contrast, non-parametric methods, particularly quantile regression, offer a more flexible solution \citep{pinson2007non, wan_2017_quantile_wind}. Quantile regression excels in providing a comprehensive quantification of uncertainty without the need to assume a predetermined functional form or distribution of forecast errors. This methodological flexibility is especially pertinent in addressing the complexities of stochastic power systems, thereby enhancing decision-making processes in critical operations such as risk management, unit commitment, economic dispatch, and optimal decision-making \citep{holttinen2012methodologies,sioshansi2010market}.
Additionally, there is growing interest in leveraging multivariate time series prediction over univariate models to better capture output dependencies. Recent studies, such as \cite{mashlakov2021assessing}, have adopted multivariate time series prediction approaches for forecasting load, wind, solar power, and price within power systems. However, these studies typically focus on zonal-level forecasting, which diverges from the generator-level analysis prevalent in most power flow problems. Furthermore, these approaches often overlook the correlations between different forecasting dimensions, a critical factor for effective scenario generation. The integration of weather forecast covariates, which can significantly influence forecasting accuracy, is also absent in their model. Another limitation of prior research is its focus on small-scale datasets and problem settings, which may not capture the complexity and scalability required for broader applications. This paper addresses these shortcomings by employing a combination of the copula method with multivariate forecasting approaches on a large-scale problem.

\subsubsection{Scenario Generation in RES}
Scenario generation in power systems is the process of creating diverse and possible future states of the system, accounting for uncertainties caused by different factors such as temperature and other weather-related variable fluctuations, which affects the power consumption and generation within the RES. Various methods have been proposed for scenario generation in power systems, including sampling-based approaches (e.g., Monte Carlo sampling, latin hypercube sampling) \citep{carmona2021glasso,9804513} and optimization-based methods (e.g., moment matching, distance matching) \citep{li2020review,9616389,lu2021case}. While these methods are foundational, they often require extensive datasets and/or are computationally demanding, particularly for generating high-dimensional scenarios. Moreover, when these methods are applied to individual dimensions (i.e., marginal distributions), they struggle to capture the critical spatio-temporal correlations present in the system.

Recently, Generative Adversarial Networks (GANs) have been employed in power systems scenario generation, notably within wind power generation, as demonstrated in \citep{ZHANG2020105388, DONG2022118387, 9351117}. These methods have been demonstrated to be effective in capturing the dynamic and complex patterns inherent in RES, thereby providing a robust framework for simulating realistic and diverse energy production scenarios. Dong et al. applies a data-driven GAN model to wind and solar scenario generation, however, weather information was not incorporated in the model \citep{DONG2022118387}. Zhang et al. utilized a point-prediction forecasting model, using GAN to learn the distribution of forecast errors and generate scenarios by adding generated residuals to the point forecasts, making the model's performance highly dependent on the accuracy of the initial forecasts \citep{ZHANG2020105388}.  Chen et al. also applied GANs to generate solar and wind scenarios, with a model that can condition on certain weather events, however does not include the weather time series such as wind speed and solar irradiation and can result in low performance \citep{8260947}. Furthermore, training GANs presents significant challenges, especially when dealing with the high-dimensional forecasting problems in RES.

The integration of the copula method with time series forecasting presents a promising solution to the aforementioned challenges by facilitating the modeling of the joint distribution. For example, the DeepVAR model utilizes a Gaussian copula to model the time-varying mutual-dependence structure, incorporating a low-rank structure on the Gaussian copula covariance to enhance computational efficiency \citep{salinas2019high}. However, DeepVAR assumes stationarity in the time series, an assumption often violated in the context of RES. An alternative strategy involves using copula methods as a post-processing step to combine predicted marginal distributions, where the covariance matrix of the Gaussian copula captures temporal dependencies \citep{golestaneh2016generation}. The application of copulas effectively bridges the gap between probabilistic forecasting models and the generation of realistic spatio-temporal scenarios. Gaussian copulas, in particular, have been extensively explored in the realm of wind power scenario generation, employing an exponential covariance function to address situations with limited historical observations \citep{ma2013scenario, bessa2016quality}. Additionally, recursive methods have been employed in wind power forecasting to refine scenario generation processes \citep{pinson2009probabilistic}.

\section{Contributions}
To address the existing gaps within high-dimensional forecasting in RES, this paper introduces a comprehensive probabilistic multivariate time series forecasting framework that leverages Gaussian copulas for enhanced forecasting and realistic scenario generation of load, and wind and solar power generation. The proposed model is augmented with both historical and forthcoming weather forecast covariates, aiming at improving forecasting and scenario generation performance across various temporal and spatial dimensions. The principal contributions of this paper are threefold:

\begin{itemize}
    \item Integration of multivariate time series forecasting with copula method for high-dimensional problem in RES to effectively capture the complex spatial and temporal dependencies. Particularly, the largest problem is a 48-hour ahead prediction for 376 solar farms, with an output dimension of 10,848.
    \item Showcasing the importance of incorporating weather forecast data, which significantly enhances the accuracy and reliability of forecasting and scenario generation processes, especially for higher prediction lead times.
    \item Comprehensive benchmarking and comparison of various time series prediction methodologies, including ARIMA, DeepAR, NLinear, DLinear, and TFT, combined with Gaussian copula. Different metrics are considered to evaluate the efficacy of these models in multi-dimensional forecasting and scenario generation within power systems.
\end{itemize}

\noindent

The analysis encompasses time series forecasting methods for the prediction and scenario generation of load, and wind and solar power generation within the Midcontinent Independent System Operator (MISO) system. In contrast with previous methods, which are mainly on small and/or low-dimensional datasets, this work considers probabilistic forecasting of a large-scale system with 6 load zones, 286 wind farms and 376 solar farms. The proposed Weather-Informed (WI) TFT model is highlighted for its superior performance across multiple evaluation metrics, in both deterministic forecasting and scenario generation. The findings of this paper showcase the comparative effectiveness of different forecasting techniques within the context of energy forecasting and uncertainty quantification, thus providing critical insights for risk-aware decision making in power system operations. 

\begin{table}[b]
\caption{Nomenclature}
    \centering
        \begin{tabular}{p{1.2cm}p{6cm}}  
        \toprule
        \multicolumn{2}{c}{\textbf{Sets}}\\ \hline
        $\mathcal{D}$ & Set of target variables  (i.e., load zones, wind farms, or solar farms)\\
        $\mathcal{D'}$ & Set of covariates \\
        $\mathcal{T}$ & Time window \\
        $\mathcal{H}$ & Forecasting horizon time window \\
        $\mathcal{W}$ & Past observation time window \\
        $\Omega$ & Dataset of input-output pairs \\\hline
        \multicolumn{2}{c}{\textbf{Indices}}\\ \hline
        $i \in \mathcal{D}$ & Spatial index \\
        $t \in \mathbb{Z}$ & Time index of forecast start times \\
        $\tau \in \mathcal{H}$ & Time index representing a specific time step within the forecast horizon \\
        \hline
        \multicolumn{2}{c}{\textbf{Variables}}\\ \hline
        $Z_{i, t}$ & Random variable for target at location $i$ and time $t$ \\
        $z_{i, t}$ & Realization of the random variable $Z_{i, t}$ \\
        $\mathbf{z}_{., t}$ & Vector realization of all joint observations in locations within $\mathcal{D}$ at time $t$ \\
        $\mathbf{z}_{i, \mathcal{T}}$ & Vector realization of observations at location $i$ within time interval $\mathcal{T}$ \\
        $\mathbf{Z}_{., \mathcal{T}}$ & Matrix of random variables for target locations within $\mathcal{D}$ over time horizon $\mathcal{T}$ \\
        $\mathbf{z}_{., \mathcal{T}}$ & Matrix of all observations in $\mathcal{D}$ over time interval $\mathcal{T}$ (i.e., realization of $\mathbf{Z}_{., \mathcal{T}}$) \\
        $\mathbf{x}_{., t}$ & Vector realization of all joint covariates within $\mathcal{D'}$ at time $t$ \\
        $\mathbf{x}_{., \mathcal{T}}$ & Matrix of all covariates in $\mathcal{D'}$ within time interval $\mathcal{T}$ \\
        $Z_i, z_i$ & Random variable and realization in target space with CDF $F_{Z_i}$ \\
        $U_i, u_i$ & Random variable and realization in uniform space $\mathcal{U}[0,1]$ \\
        $V_i, v_i$ & Random variable and realization in Gaussian space $\mathcal{N}(0,1)$ \\
        $F_{i, \tau} (z_{i, \tau})$ & Marginal CDF for target at location $i\in \mathcal{D}$ \\ & at time $\tau \in \mathcal{H}$ \\
        $\mathbf{F}_{., \mathcal{H}}(\mathbf{z}_{., \mathcal{H}})$ & Joint CDF for all targets within $\mathcal{D}$ over \\& time window $\mathcal{H}$ \\
        $\hat{F}, \hat{\mathbf{F}}$ & Empirical marginal and joint CDFs \\
        $\mathbf{0}, \mathbf{1}$ & Vectors of all zeros and ones, respectively \\
        $\mathbf{R}$ & Correlation matrix \\
        $\mathbf{C}$ & Copula function \\
        $\Phi$ & Standard normal CDF \\
        $\mathbf{\Phi}_{\mathbf{R}}$ & Multivariate normal CDF with standard normal marginal CDFs and correlation matrix $\mathbf{R}$ \\\hline
        \multicolumn{2}{c}{\textbf{Operators}}\\ \hline
        $\text{vec}$ & Matrix vectorization \\
        max & Largest value operator \\
        \bottomrule
    \end{tabular}
\end{table}

\section{Problem Formulation}
\label{section: material and methods}

\begin{figure*}[b]
    \centering
    \subfloat[]{
        \includegraphics[trim=5cm 7cm 7cm 5cm, clip, width=0.7\linewidth]{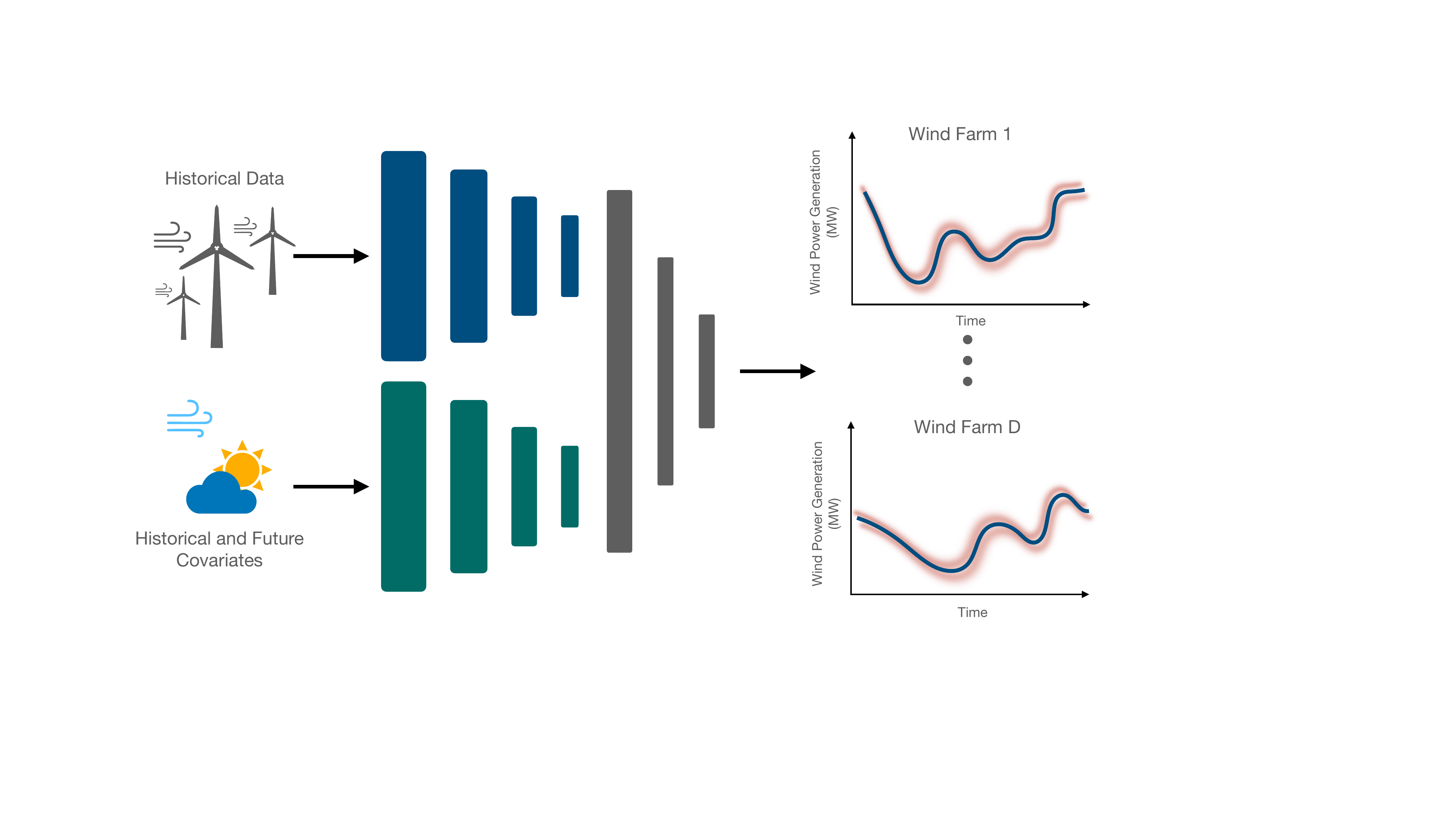}
    }
    \hfill
    \subfloat[]{
        \includegraphics[trim=0cm 0.1cm 0cm 0cm, clip, width=1\linewidth]{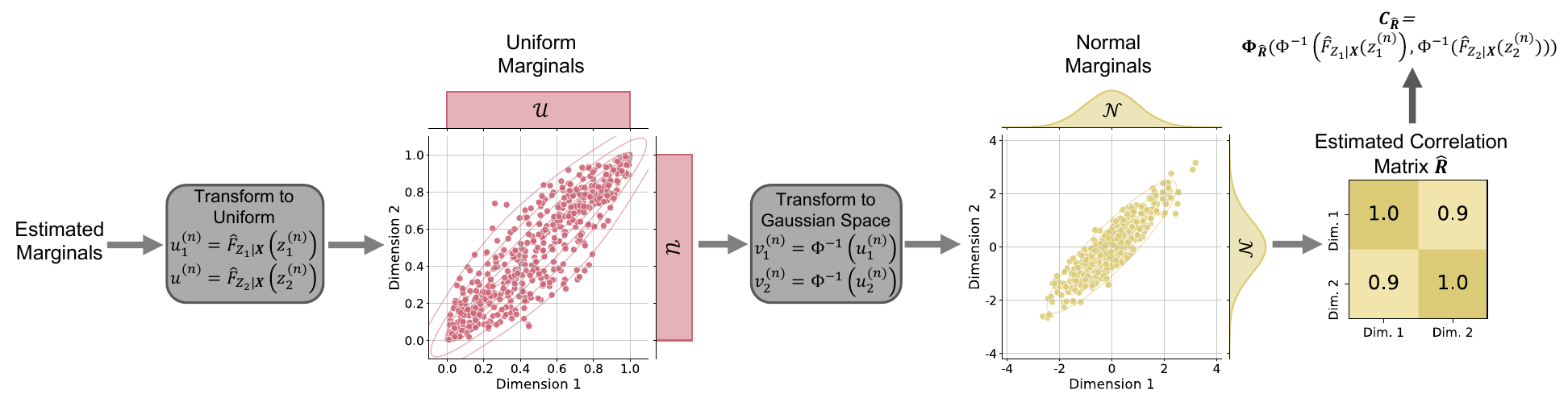}
    }

    \hfill
    \subfloat[]{
        \includegraphics[trim=0cm 0.1cm 0cm 0cm, clip, width=1\linewidth]{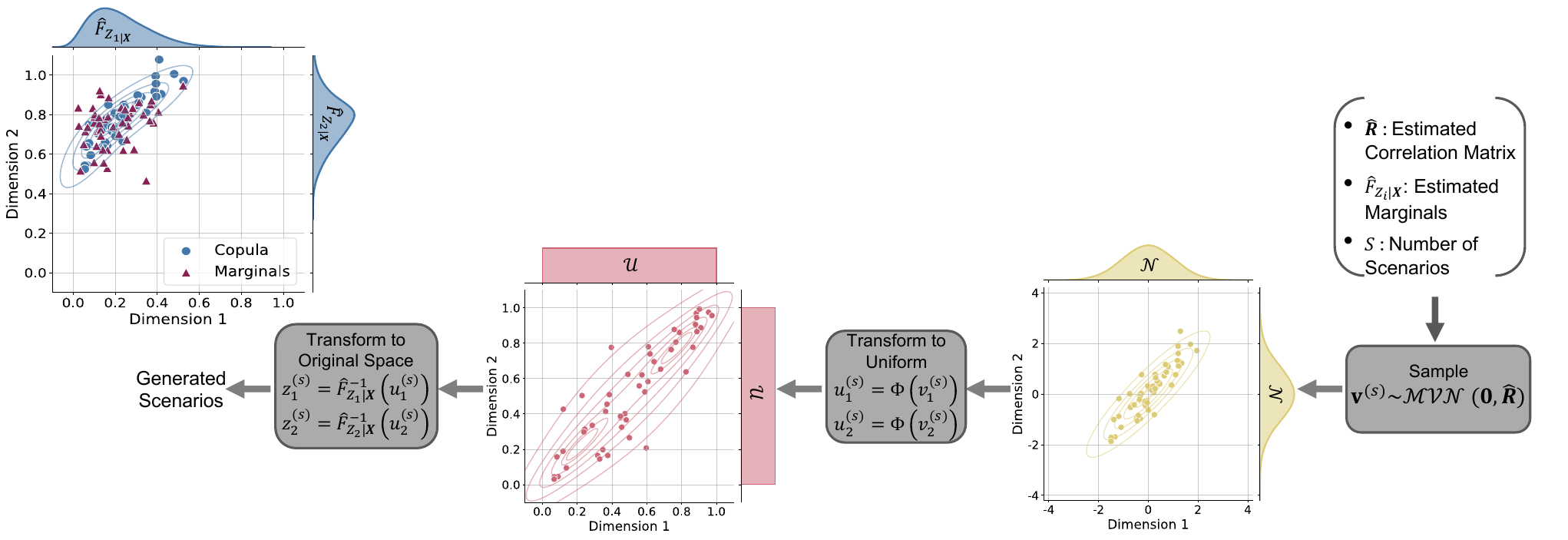}
    }
    \caption{Schematic of the proposed method for scenario generation in RES. (a) Schematic of the weather informed multivariate probabilistic forecasting for wind farms, given the historical wind power generation and the past and future weather covariates. (b) Schematic of estimating the Gaussian copula for a 2D example. (c) Schematic of the scenario generation steps given the estimated copula and estimated marginals for a 2D problem. Examples of generated scenarios using marginals are also presented in (c) with red triangles.}
    \label{fig:flow_chart}
\end{figure*} 

This section outlines the problem formulation and introduces the notations used for the probabilistic multivariate time series forecasting of load, and wind and solar power generation, that incorporates weather data.

\subsection{Problem Setting}
Let $Z_{i, t}$ denote a random variable corresponding to the target $i \in \mathcal{D} = \{1, .., D\}$ (i.e., a load, a wind farm, or a solar farm) at time $t \in \mathbb{Z}$, and denote its realization as $z_{i, t} \in \mathbb{R}$. The vector realization of all joint observations at time $t$ is defined as:
\begin{align}
    \mathbf{z}_{., t} = (z_{1, t}, ..., z_{D, t}) \in \mathbb{R}^D
\end{align}

For instance, in the context of wind farms, $z_{i,t}$ represents the output generated power of wind farm $i$ at time $t$, measured in megawatts (MW), and $\mathbf{z_{., t}}$ is the vector representing output of all wind farms at time $t$. Similarly, for an individual target $i$, the realization of the univariate time series for time interval $\mathcal{T} = \{t+1, ..., t+T \} \subseteq \mathbb{Z}$ is denoted by:
\begin{align}
    \mathbf{z}_{i, \mathcal{T}} = (z_{i, t+1}, ..., z_{i, t+T}) \in \mathbb{R}^{T}
\end{align}

\noindent
Hence, the multivariate time series realization corresponding to the targets in the system from time $t+1$ to $t+T$ is denoted as:
\begin{align}
\label{eq: multi time series}
    \mathbf{z}_{\cdot, \mathcal{T}} &= 
    \begin{pmatrix}
        \text{---} & \mathbf{z}_{1, t+1:t+T} & \text{---} \\
            & \vdots & \\
        \text{---} & \mathbf{z}_{D, t+1:t+T} & \text{---}
    \end{pmatrix}
    \in \mathbb{R}^{D \times T}
\end{align}

\noindent
The paper also considers so-called \textit{covariates}, denoted by $\mathbf{x}_{.,t}$ with support in $\mathbb{R}^{D'}$. Thereby, for $j \in \mathcal{D'}=\{1, ..., D' \}$, $x_{j, t}$ represents the observed value of the $j$-th covariate at time $t$, and the vector representation of all covariates at time $t$ is denoted by $\mathbf{x}_{., t} = (x_{1, t}, ..., x_{D', t})$. The corresponding covariates in the paper include time encoding, e.g., day of the week, and the weather-related quantities such as temperature, humidity, cloud coverage, wind speed, etc. Hence, the matrix realization of the covariates is denoted as below:
\begin{align}
    \mathbf{x}_{\cdot, \mathcal{T}} &= 
    \begin{pmatrix}
        \text{---} & \mathbf{x}_{1, t+1:t+T} & \text{---} \\
            & \vdots & \\
        \text{---} & \mathbf{x}_{D', t+1:t+T} & \text{---}
    \end{pmatrix}
    \in \mathbb{R}^{D' \times T}
\end{align}

\noindent
The details corresponding to each target and the covariates are discussed in Section \ref{sec: experimental setup}.

\section{Multivariate Forecasting} \label{sec: prob_formulation prob_forecasting}
Let $t\in \mathbb{Z}$ and consider the forecasting horizon $\mathcal{H} = \{t+1, ..., t+H\}$ of length $H$, and the past observation window of length $W$ as $\mathcal{W} = \{t-W+1, .., t-1, t\}$. A time series forecasting model predicts the target $\mathbf{z}_{.,\mathcal{H}} \in \mathbb{R}^{D \times H}$ as $\mathbf{\hat{z}}_{.,\mathcal{H}} \in \mathbb{R}^{D \times H}$. Figure \ref{fig:flow_chart}(a) shows schematic of the DL-based model for multivariate probabilistic forecasting of the wind power generation given historical data and covariates.
More generally, probabilistic forecasting can be considered as modeling the joint cumulative distribution function (CDF) of the target variables conditioned on the past observations $\mathbf{z}_{., \mathcal{W}} \in \mathbb{R}^{D \times W}$, past covariates $\mathbf{x}_{., \mathcal{W}} \in \mathbb{R}^{D' \times W}$, and the future covariates $\mathbf{x}_{., \mathcal{H}} \in \mathbb{R}^{D' \times H}$ as 
\begin{align}
    \label{eq: joint_distribution}
        \mathbf{F}_{\mathcal{H}}(\textbf{z}_{., \mathcal{H}}) &= 
        \mathbb{P}\left(
            \mathbf{Z}_{., \mathcal{H}} \leq \mathbf{z}_{., \mathcal{H}}
            \ \middle| \ 
            \mathbf{z}_{., \mathcal{W}},
            \mathbf{x}_{., \mathcal{W}},
            \mathbf{x}_{., \mathcal{H}}
        \right)
\end{align}

\noindent
Modeling this joint CDF directly is challenging due to the high-dimensionality of the problems at hand. 
An alternative, in the multivariate forecasting setting, is to jointly predict the following marginal distributions:
\begin{align}
    \label{eq: marginal_distribution}
    F_{i, \tau}(z_{i, \tau}) = \mathbb{P} (Z_{i, \tau} \leq z_{i, \tau} \mid \mathbf{z}_{., \mathcal{W}}, \mathbf{x}_{., \mathcal{W}}, \mathbf{x}_{., \mathcal{H}}),
\end{align}
where $i \in \mathcal{D}$ is the spatial index indicating the location of the target and $\tau \in \mathcal{H}$ represents the time index within the forecasting horizon. In other words, the multivariate forecasting model takes as input the historical data and covariates, which are shared across different targets, and predicts the marginal distributions. Then, given the estimated marginals as the outputs of the multivariate forecasting, the next critical step is to restore the spatial and temporal correlations among the target variables. This is necessary for realistic scenario generation, particularly within the RES where the targets (e.g., electricity demand and outputs of multiple wind farms or solar farms at different locations and time horizons) are highly correlated. This paper restores the joint distribution of the random variable $\mathbf{Z}_{., \mathcal{H}}$ using a copula method, as discussed in Section \ref{sec: copula}.

\section{Restoring Correlations via Copula}\label{sec: copula}

The copula method provides a robust statistical tool for modeling the underlying dependencies and correlations in high-dimensional probabilistic forecasting. This method enables the estimation of the joint distribution by combining the marginal distributions with a copula function. In the context of RES, the copula method facilitates the generation of realistic scenarios that accurately reflect the true variability and spatial-temporal interdependencies.

\subsection{Copula Method}

A function $\mathbf{C}: [0, 1]^d \rightarrow [0, 1]$ is a \emph{copula} if it is the joint CDF of a $d$-dimensional random vector $\mathbf{U} = (U_1, \ldots, U_d)$ with uniformly distributed marginals on the interval $[0, 1]$.
Sklar's Theorem \citep{sklar1959fonctions} shows that any multivariate CDF can be expressed as the product of its marginal distributions and a copula function.

\begin{theorem}[Sklar's Theorem ]
Let $\mathbf{Z}$ be a $d$-dimensional random variable with joint CDF $\mathbf{F}_{\mathbf{Z}}$ and marginal CDFs $F_{Z_1}, \ldots, F_{Z_d}$.
There exists a copula $\mathbf{C}$ such that
\begin{align}
    \mathbf{F}_{\mathbf{Z}}(z_{1}, ..., z_{d}) = \mathbf{C}\left(F_{Z_1}(z_1),...,F_{Z_d}(z_d) \right),
    \label{eq: sklar}
\end{align}
and the random variables $U_i = F_{Z_i}(Z_i)$ are uniformly distributed on interval $[0, 1]$ (i.e., $U_i \sim \mathcal{U} [0, 1]$).
The copula $\mathbf{C}$ is unique, given \textit{continuous} marginals $F_{Z_1}, \ldots, F_{Z_d}$.
\end{theorem}

Sklar's theorem thus enables the decomposition of any multivariate CDF into its marginal CDFs and a copula, without specific assumptions on the marginals. Hence, given the estimated marginal CDFs, it is possible, via a Copula function, to reconstruct the joint CDF and represent the underlying dependencies and correlations among different dimensions. This is particularly crucial in probabilistic forecasting in RES where spatial and temporal correlations should be restored in high dimensions.

\subsection{Conditional Copula}
Sklar's theorem can be extended to modeling conditional distributions, which is particularly important in regression and time series forecasting \citep{patton2006modelling, patton2009copula}. Given a target random vector $\mathbf{Z} = (Z_1, \ldots, Z_d)$ conditioned on a set of input variables $\mathbf{X}$, the conditional copula aims at modeling $\mathbf{F}_{\mathbf{Z}|\mathbf{X}}$ from its marginals $F_{Z_1 | \mathbf{X}}, \ldots, F_{Z_d | \mathbf{X}}$ as follows:
\begin{equation}
    \begin{split}
        \mathbf{F}_{\mathbf{Z}|\mathbf{X}} = \mathbf{C}(F_{Z_1 | \mathbf{X}}(z_1|\mathbf{x}), \ldots, F_{Z_d| \mathbf{X}}(z_d|\mathbf{x}))
    \end{split}
\end{equation}
It is important to note that, in general, the input variables $\mathbf{X}$ should be consistent across different marginals and the copula. For further details and considerations regarding different input variable sets under specific assumptions, readers are referred to \citep{fermanian2012time}.

\subsection{Gaussian Copula} \label{sec: gaussian copula}
One of the most popular and effective ways of linking the joint distribution into its marginals is the Gaussian copula. Considering $\Phi$ as the CDF of the standard normal distribution and $\mathbf{\Phi_R}$ as the multivariate normal CDF with mean vector of $\mathbf{0}$ and covariance matrix of $\mathbf{R}$, the Gaussian copula $\mathbf{C_R} : [0, 1]^d \rightarrow [0, 1]$ is defined as:
\begin{equation}
    \begin{split}
        \mathbf{C_R} = \mathbf{\Phi_R} \bigl(\Phi^{-1}(u_1), \ldots, \Phi^{-1}(u_d)\bigr)
    \end{split}
\end{equation}
As a result, the multivariate joint CDF $\mathbf{F}_{\mathbf{Z}|\mathbf{X}}$ can be decomposed as:
\begin{equation}
    \begin{split}
        \mathbf{F}_{\mathbf{Z}|\mathbf{X}} = \\ & \mathbf{\Phi_R} \Bigl(\Phi^{-1} \bigl(F_{Z_1|\mathbf{X}}(z_1|\mathbf{x})\bigr), \ldots, \Phi^{-1}\bigl(F_{Z_d | \mathbf{X}}(z_d|\mathbf{x})\bigr)\Bigr)
    \end{split}
    \label{eq:copula-joint-dist}
\end{equation}
The two main pillars to estimate the joint CDF under the Gaussian copula framework are: 1) the marginal CDFs $F_{Z_i|\mathbf{X}}$ for $i\in \{1, \ldots, d\}$ and 2) the correlation matrix $\mathbf{R} \in \mathbb{R}^{d \times d}$, which is responsible for modeling the correlations among different dimensions. 

Figure \ref{fig:flow_chart}(b) shows the schematic of the approach for estimating the Gaussian copula for a 2D (i.e., $d = 2$) problem. First, the marginal CDFs are estimated as $\hat{F}_{Z_i|\mathbf{X}}$ using a probabilistic forecasting model and a training dataset $\Omega_\text{train}$ with $N = |\Omega_\text{train}|$ samples.
Then, given the training covariate and target pairs (i.e., $(\mathbf{x}^{(n)}, \mathbf{z}^{(n)}) \in \Omega_\text{train}, n\in\{1, \ldots, N\}$), and the estimated marginals $\hat{F}_{Z_i|\mathbf{X}}$, each of the target dimensions $z_i^{(n)}$ is transformed into $u_i^{(n)} = \hat{F}_{Z_i|\mathbf{X}}$. The transformed variables $u_i^{(n)}$ are distributed uniformly on $[0, 1]$ if the underlying model that estimates marginals is well-calibrated (refer to the probability integral transform in the Appendix for more details). Following this step, the corresponding data is transformed through the inverse of the standard normal CDF to $v_i^{(n)}=\Phi^{-1}(u_i^{(n)})$. This results in a space with standard normal marginal CDFs. Finally, the correlation matrix is estimated as $\mathbf{\hat{R}}$, using Pearson correlation over $\mathbf{v}^{(n)} = (v_1^{(n)}, \ldots, v_d^{(n)}), n\in \{1, \ldots, N\}$. The estimated correlation matrix is then used along with the estimated marginals to form the Gaussian copula as presented in Equation \eqref{eq:copula-joint-dist}. In short, given a dataset with dimension $d \times N$, where $d$ corresponds to the dimension of the copula and $N$ is the number of training data points, i.e., $N= |\Omega_{train}|$, the correlation matrix is estimated as:

\begin{equation}
\label{eq: pearson's correlation}
\hat{\mathbf{R}} = \text{corr} \left( \left[ 
\begin{array}{c}
\text{--- } \Phi^{-1} \left( \hat{F}_{z_{1} \mid \mathbf{x}} \left( z_1^{(n)} \mid \mathbf{x}^{(n)} \right) \right) \text{---} \\
\vdots \\
\text{--- } \Phi^{-1} \left( \hat{F}_{z_d \mid \mathbf{x}} \left( z_d^{(n)} \mid \mathbf{x}^{(n)} \right) \right) \text{---}
\end{array}
\right]_{d \times N} \right)
\end{equation}

\subsection{Scenario Generation}
\label{sec:scenario generation}
The combination of the multivariate probabilistic forecasting and copula method provides the estimated joint distribution of the target variables, incorporating the correlations among different dimensions. Consequently, it is possible to generate probabilistic scenarios by sampling from the estimated joint distribution that captures the underlying correlations between different dimensions.

The scenario generation steps are outlined in Algorithm \ref{alg:pseudo-code general} and are depicted in Figure \ref{fig:flow_chart}(c) for a 2D (i.e., $d = 2$) example. The algorithm takes as input the estimated correlation matrix $\mathbf{\hat{R}}$, the estimated marginals $\hat{F}_{Z_i|\mathbf{X}}$ for $i \in \{1, \ldots, d\}$, and the number of required scenarios (i.e., the number of samples from the joint distribution) $S$, and it outputs $\mathbf{\hat{z}}^{(s)} = (\hat{z}_1^{(s)}, \ldots, \hat{z}_d^{(s)})$ for $s \in \{1, ..., S \}$. The algorithm starts by sampling $\mathbf{v}^{(s)} = (v_1^{(s)}, \ldots, v_d^{(s)})$ from a multivariate normal distribution with $\mathbf{0}$ mean and estimated covariance matrix $\mathbf{\hat{R}}$ (i.e., $\mathcal{MVN}(\mathbf{0}, \mathbf{\hat{R}})$). The samples are then transformed using the CDF of the standard normal into $u_i^{(s)} = \Phi(v_i^{(s)})$, which has uniform marginals (as shown in Figure \ref{fig:flow_chart}(c)). Finally, the samples are mapped into the original space via $\hat{z}^{(s)}_i = \hat{F}^{-1}_{Z_i|\mathbf{X}}(u_i^{(s)})$. The resulting vector $\mathbf{\hat{z}}^{(s)}$ corresponds to the samples generated using the copula method, which captures the interdependencies and correlations among the different dimensions.

\begin{table}[t]

\centering
\resizebox{0.9\linewidth}{!}{
\footnotesize{
\begin{minipage}{\linewidth}
\begin{algorithm}[H]
\caption{Scenario Generation Pseudo-code}
\label{alg:pseudo-code general}
\begin{algorithmic}
\State \textbf{Inputs:} $\mathbf{\hat{R}}$ (Estimated Correlation Matrix), \\  \hspace{3 em} $\hat{F}_{Z_i|\mathbf{X}}$ (Estimated Marginals for $i \in \{1, \ldots, d\}$), \\ \hspace{3 em}  $S$ (Total Number of Required Scenarios)
\State \textbf{Output:} Generated Scenarios: $\mathbf{\hat{z}}^{(s)} = (\hat{z}_1^{(s)}, \ldots, \hat{z}_d^{(s)})$ for $s \in \{1, \ldots, S\}$

\For{$s = 1$ to $S$} \Comment{Can be done in parallel}
    \State Sample $\mathbf{v}^{(s)} = (v_1^{(s)}, \ldots, v_d^{(s)})$ from $\mathcal{MVN}(\mathbf{0}, \mathbf{\hat{R}})$
    \For{each $i \in \{1, \ldots, d\}$}
        \State Compute $u_i^{(s)} = \Phi(v_i^{(s)})$
        \State Map to the original space: \\  \hspace{3 em} $\hat{z}_i^{(s)} = \hat{F}_{Z_i|\mathbf{X}}^{-1}(u_i^{(s)})$
    \EndFor
\EndFor
\end{algorithmic}
\end{algorithm}
\end{minipage}
}
}
\end{table}

\subsection{Copula for Time Series Scenario Generation}

This section extends the Gaussian copula method to the context of time series forecasting in RES, where the objective is to model a joint CDF for all sites and time steps, and generate scenarios that capture both spatial and temporal dependencies across the forecasting horizon. In this setting, the joint conditional CDF $\mathbf{F}_{\mathcal{H}}(\mathbf{z}_{., \mathcal{H}})$ is over $d = D\times H$ dimensions (recall that $D = |\mathcal{D}|$ is the number of locations and $H = |\mathcal{H}|$ is the number of time steps in the forecast horizon). Similar to what is presented in Section \ref{sec: gaussian copula}, the joint conditional CDF can be derived from: 1) the estimated conditional marginals $\hat{F}_{i, \tau}(z_{i, \tau})$ (where $i \in \mathcal{D}$ and $\tau \in \mathcal{H}$) and 2) a Gaussian copula with an estimated correlation matrix $\mathbf{\hat{R}}\in \mathbb{R}^ {(D \times H) \times (D \times H)}$ that captures the spatio-temporal correlations as follows:

{\small
\setlength{\mathindent}{0pt} %
\begin{align}
\label{eq:copula-time series}
    \mathbf{\hat{F}}_{\mathcal{H}}(\mathbf{z}_{., \mathcal{H}}) & = \mathbf{\Phi}_\mathbf{\hat{R}} \Bigl( \Phi^{-1} \bigl(\hat{F}_{1, t+1} (z_{1, t+1}) \bigr), \dots , \Phi^{-1} \bigl( \hat{F}_{1, t+H} (z_{1, t+H})\bigr), \nonumber\\ 
    &  \quad \quad \quad \quad \quad\quad\quad \quad \quad \quad \quad \dots,\\
    &\quad \quad \quad \Phi^{-1} \bigl( \hat{F}_{D,t+ 1} (z_{D, t + 1})\bigr) \dots , \Phi^{-1} \bigl(\hat{F}_{D, t + H} (z_{D, t + H})\bigr) \Bigr) \nonumber
\end{align}
}
\noindent
Note that $\mathbf{\hat{F}}_{\mathcal{H}}$ and $\hat{F}_{i,\tau}$ (defined in \eqref{eq: joint_distribution} and \eqref{eq: marginal_distribution}, respectively) are conditioned on the inputs 
$\mathbf{z}_{., \mathcal{W}}$,
$\mathbf{x}_{., \mathcal{W}}$,
and $\mathbf{x}_{., \mathcal{H}}$, which are omitted in Equation \eqref{eq:copula-time series} for simplicity. Also, $\tau$ depends on $t$, which corresponds to the starting of the forecasting time window. The marginals are estimated using multivariate probabilistic time series forecasting given a training dataset and the corresponding covariates as:

{\small
\setlength{\mathindent}{0pt} %
\begin{equation*}
\Omega_\text{train} = \left\{ 
\left( 
\{\mathbf{z}_{., \mathcal{W}}^{(t)}, \mathbf{x}_{., \mathcal{W}}^{(t)}, \mathbf{x}_{., \mathcal{H}}^{(t)}\}, \mathbf{z}_{., \mathcal{H}}^{(t)} 
\right) 
\middle| 
\begin{aligned}
    t \in & \{W, W+1, \\ & \quad \quad \ldots, N-H\}, \\
    \mathcal{H} = & \{t + 1,\dots,t+H\}, \\
    \mathcal{W} = & \{t-W+1,\dots,t\}
\end{aligned}
\right\},
\end{equation*}} 
where $N$ is the total length of the training time series dataset and $t$ is the time index in the rolling time window.

Next, using the estimated marginal CDFs and following the steps outlined in Section \ref{sec: gaussian copula}, the correlation matrix $\hat{\mathbf{R}} \in \mathbb{R}^ {(D \times H) \times (D \times H)}$ can be estimated using Pearson correlation, with $N' = N-W-H+1$ samples (i.e., the training dataset excluding the few first and last time steps).

The scenario generation follows the process outlined in Section \ref{sec:scenario generation} and Algorithm 1. Samples are drawn from $\mathcal{MVN}(\mathbf{0}, \hat{\mathbf{R}})$, producing $S$ vectors with dimensions $1 \times (D\times H)$. Each sample is then reshaped into a matrix of size $D \times H$, where the rows and the columns correspond to locations and time, respectively.

\section{Experiment Setup} \label{sec: experimental setup}
A comprehensive performance evaluation was conducted with large-scale data using various time series forecasting methods. Both deterministic and probabilistic forecasting methods are employed to assess their efficacy under different metrics. Additionally, the experiments incorporate weather information to evaluate its importance in predicting RES. The forecast horizon $\mathcal{H}$ is set to 48 hours, which aligns with the common practice for day-ahead market-clearing and reliability assessment in power systems \citep{werho2021scenario}. However, the methodology can be extended to longer or shorter time horizons depending on the specific requirements.

    \begin{table}[h]
        \centering
        \caption{Dataset Summary: This table presents the overall minimum and maximum values across all time series. For wind and solar, the maximum values represent the capacity of the largest wind or solar farms. For the load time series, which does not have a predefined capacity, the capacity is defined by the highest observation in the training dataset. The minimum value for each type corresponds to the lowest recorded observation across all time series.}
        \label{tab:DataSummary}
        \resizebox{\linewidth}{!}{%
        \begin{tabular}{rrrrrr} \toprule
            Type  & $D$ & Level & Total(GW) & Min (MW) & Max (MW) \\ \hline 
                 load  & 6              & zone  &     124.2   &     9424.6     &     32148.5     \\
                                       wind  & 286            & unit  & 21.4          &   0.6       &   495.0       \\
                                       solar & 376            & unit  &    1.8    &  0.2        &   138.0       \\
            \bottomrule
        \end{tabular}%
        }
    \end{table}

\subsection{Dataset}

Table \ref{tab:DataSummary} provides summary statistics of the MISO dataset that is used in this study which contains time series data for the years 2018 and 2019 with hourly resolution within the MISO system \citep{nrelperformdata}. This dataset includes zone-level time series for loads and unit-level time series of wind and solar power generation. Additionally, it includes forecasts generated by NREL's System Advisor Model (SAM) for the year 2019, which is considered for comparison with the developed models in this study.

\begin{table*}[!Hb]
\centering
\caption{Description of Meteorological Variables \citep{ecmwf_tigge_2024}.}
\label{tab:variables}
\begin{tabularx}{\linewidth}{lX}
\hline
Variable & Description \\
\hline
10u & - Eastward component of the 10 metre wind (\textit{U-component}): Measures the horizontal speed of air moving towards the east at a height of ten metres above the Earth's surface, in metres per second. \\
10v & - Northward component of the 10 metre wind (\textit{V-component}): Measures the horizontal speed of air moving towards the north at a height of ten metres above the Earth's surface, in metres per second. \\
2t  & - 2 metre temperature: Temperature of the air measured at 2 metres above the surface of the Earth, in kelvins (K). \\
skt & - Skin temperature: Temperature at the interface between the Earth and the atmosphere, in kelvins (K). Measured directly at the Earth's surface. \\
sp  & - Surface pressure: Atmospheric pressure exerted on the Earth's surface by the air above, measured in pascals (Pa). Reflects the weight of the air vertically above a point. \\
ssr & - Net surface solar radiation: Total solar radiation (shortwave radiation) reaching a horizontal plane at the Earth's surface, adjusted for the albedo, measured in joules per square metre (J/m$^2$). \\
tcc & - Total cloud cover: Proportion of the sky occluded by clouds, measured on a scale from 0 (no cover) to 1 (complete cover). \\
\hline
\end{tabularx}
\end{table*}

To explore the effect of weather data, this study incorporates an additional dataset of weather forecasts for the MISO system. The weather forecasts are gathered from the publicly available datasets produced by the European Center for Medium-Range Weather Forecasts (ECMWF), particularly from the THORPEX Interactive Grand Global Ensemble (TIGGE) dataset \citep{ecmwf, ecmwf_tigge_2024}. This dataset provides essential meteorological variables including surface pressure, wind speed, cloud cover, skin temperature, solar irradiation, as outlined in Table \ref{tab:variables}. Issued twice everyday at 12:00 and 00:00 UTC, the forecasts from the TIGGE database are provided in 6-hour intervals \citep{ecmwf_tigge_2024}. The forecast data is collected on a 1° latitude by 1° longitude grid covering the MISO territory. The 6-hour temporal resolution time series from the ECMWF are then interpolated to hourly granularity to align with the load, and wind and solar power time series data used in this study.

\subsection{Methods} \label{sec: models}

In this study, different methods are employed to determine their efficacy for 48-hour forecasting horizon prediction in RES and load. The models encompass both traditional statistical methods and advanced DL-based techniques with and without incorporation of weather data, to provide a comprehensive performance comparison. The models include:

\begin{itemize}
    \item ARIMA: This model is implemented in a univariate framework due to the shortcoming of the autoregresive methods in handling high-dimensional input data \citep{arima}. 

    \item Dlinear: This is a linear model tailored for time series forecasting. The implementation in this work considers the spatio-temporal correlations by regressing the forecasts over the previous observations and past and future covariates \citep{zeng2022transformers}.

    \item Nlinear: This model normalizes each input by subtracting the last value in the sequence and then adds
    it back after passing through the linear layer to address distribution shift. In this paper, the model is adapted to probabilistic setting by using quantile regression objective function \citep{zeng2022transformers}.

    \item DeepAR: This is an RNN-based model designed to capture the temporal dependencies and provide multi-step ahead forecast by recursively applying the model, using its own predictions as inputs for subsequent steps. In this work, the model's output parameters are parametrized using a Beta distribution to account for the capacity factors in RES, which are inherently bounded between 0 and 1 \citep{salinas2020deepar}.

    \item TFT: This is a transformer-based model designed for multi-horizon forecasting with mixed-type inputs, including static covariates, known future inputs, and past exogenous time series. Quantile loss is utilized for training the model in this work \citep{LIM20211748}.

\end{itemize}

\noindent
The models are trained using historical data of the targets from 2018 within the MISO system, with the remaining data reserved for testing. At the beginning of each month, a new model is trained using all available observations since Jan. 1st, 2018 up to midnight on the first day of that month. The first week of each month is designated for validation purposes (i.e., for fine tuning the model parameters), and the models are subsequently evaluated during the second week of each month. The DL models are trained both with and without weather data to assess the importance of weather covariates. Evaluations are conducted only for the first nine months, as the ECMWF data was not available from October 2019 onward.

Weather features are utilized as additional covariates in the forecasting tasks for the WI-based models. These covariates include both historical weather data and weather forecasts, which are available in advance as discussed in the previous section. For each target variable, weather information from the closest region on a 1° by 1° grid is used. For load forecasting, skin temperature and total cloud cover are incorporated as key variables. For wind power generation forecasting, the model uses wind speed, surface pressure, and both V and U wind components, which indicate horizontal airflow towards the north and east, respectively. In forecasting solar power generation, total cloud cover and surface solar radiation are key variables. When generating forecasts, the actual weather information from the past and weather forecasts for the upcoming forecasting horizon are fed into the WI-based models..

The ARIMA model parameters are optimized using the open-source Statsforecasts package \citep{nixtlaverse_autoarima}, which features an AutoARIMA function for parameter selection. Training of the DL-based models are performed in Pytorch using Adam optimizer \citep{kingma2014adam}. Early stopping with patience of 10 steps in conjunction with a learning rate scheduler (ReduceLROnPlateau) with patience of 5 steps is used for all models. The remaining hyperparameters for each model are optimized using Optuna \citep{optuna_2019}. The details about the training process are provided in \ref{appendix:sec:trainingprocess}. 

\begin{table*}[h]
    \centering
     \caption{Summary and Description of Evaluation Metrics. \\
    \footnotesize{ $^*$ $\hat{\text{ED}}(\mathbf{x} , \mathbf{y}) = \frac{2}{ S} \sum_{i=1}^{S} \| \mathbf{x}_{i} - \mathbf{y} \|  
    - \frac{1}{S^{2}} \sum_{i=1}^{S} \sum_{j=1}^{S} \| x_{i} - x_{j} \|,$ more details in appendix. } \\
    \footnotesize{$^{**}\operatorname{max}\{\mathbf{z}_{i,\mathcal{T}_{\text{train}}}\}$ refers to the capacity for wind/solar farm $i$, which is the maximum observation in the training dataset for time series $i$.}
    }
    \begin{tabular}{>{\centering\arraybackslash}m{1.4cm}>{\centering\arraybackslash}m{6cm}>{\arraybackslash}m{8cm}}
        \hline
        Metric & Formula &  Description \\
        \hline
        $\text{NMAE}_{\ind}^{**}$& 
        $\frac{1}{N_{\text{test}}DT} \sum\limits_{t=1}^{N_{\text{test}}} \sum\limits_{i \in \mathcal{D}, \tau \in \mathcal{H}} \left|\frac{\hat{z}_{i,\tau} - z_{i,\tau}}{\operatorname{max}\{\mathbf{z}_{i,\mathcal{T}_{\text{train}}}\}}\right|$ & Computed on individual space-time dimensions, normalized by generator capacity or maximum value in the first year of training dataset, measuring model's accuracy relative to the scale of the time series. \\
        
        \hline
        $\text{NMAE}_{\mys}$ & 
        $\frac{1}{N_{\text{test}}T} \sum\limits_{t=1}^{N_{\text{test}}} \sum\limits_{\tau \in \mathcal{H}} \left|\frac{\sum\limits_{i=1}^{D} \hat{z}_{i,\tau} - \sum\limits_{i=1}^{D} z_{i,\tau}}{\sum\limits_{i=1}^{D} \operatorname{max}\{\mathbf{z}_{i,\mathcal{T}_{\text{train}}}\} }\right|$ & Computed on the summation over space, measuring model's accuracy relative to the scale of the time series. \\
        \hline
        $\text{RMSE}_{\ind}$ & 
        $\sqrt{\frac{1}{N_{\text{test}}DT} \sum\limits_{t=1}^{N_{\text{test}}}  \sum\limits_{i \in \mathcal{D}, \tau \in \mathcal{H}} \left(\hat{z}_{i,\tau} - z_{i,\tau}\right)^2}$ & Root Mean Squared Error computed on individual level over all space and time dimensions, measuring model's accuracy in MW scale. \\
        \hline
        $\text{RMSE}_{\mys}$ & 
        $\sqrt{\frac{1}{N_{\text{test}}T} \sum\limits_{t=1}^{N_{\text{test}}} \sum\limits_{\tau \in \mathcal{H}} \left(\sum_{i=1}^{D} \hat{z}_{i,\tau} - \sum\limits_{i=1}^{D} z_{i,\tau}\right)^2}$ & Root Mean Squared Error computed on the summation of wind power, solar power, or load over space, measuring model's accuracy in MW scale. \\
        \hline
        $\text{ED}_{\ind}^*$ & 
        $\frac{1}{N_{\text{test}}} \sum\limits_{t=1}^{N_{\text{test}}}\hat{\text{ED}}(\{\operatorname{vec}(\hat z_{\cdot,\mathcal{H}})^{(s)}, s\in \mathcal{S}\},\operatorname{vec}(z_{\cdot,\mathcal{H}}) )$ & Average energy distance between the actual and probabilistic forecasts, capturing the joint distribution over space and time. \\
        \hline
        $\text{ED}_{\mys}^*$ & 
        $\frac{1}{N_{\text{test}}} \sum\limits_{t=1}^{N_{\text{test}}} \hat{\text{ED}}(\{\hat z_{\cdot,\mathcal{H}}^{(s)}, s\in \mathcal{S})\}, \mathbf{1}^{\rm T} z_{\cdot,\mathcal{H}} )$ & Energy distance computed on space-sum level, capturing the joint distribution over time on the space-sum level. \\
        \hline
        $\text{VS}_{\mys}$ & 
        \hspace{0.8cm}\parbox{5cm}{$\frac{1}{N_{\text{test}}}\sum\limits_{t=1}^{N_{\text{test}}}\sum\limits_{\tau_{1} \in \mathcal{H}} \sum\limits_{\tau_{2} \in \mathcal{H}} \left( \left| \sum\limits_{i=1}^{D} z_{i, \tau_{1}} - \sum\limits_{i=1}^{D} z_{i, \tau_{2}}\right|^{\frac{1}{2}} \right.$\\
        $- \left. \frac{1}{S} \sum\limits_{s=1}^{S} \left| \sum\limits_{i=1}^{D} \hat{z}_{i, \tau_{1}}^{(s)} - \sum\limits_{i=1}^{D} \hat{z}_{i, \tau_{2}}^{(s)} \right|^{\frac{1}{2}} \right)$} & Evaluated for summation of load and renewable generation predictions over space, capturing temporal correlations. \\
        \hline
        $\text{VS}_{\myt}$ & 
        \hspace{0.8cm}\parbox{5cm}{$\frac{1}{N_{\text{test}}}\sum\limits_{t=1}^{N_{\text{test}}}\sum\limits_{i_{1} \in \mathcal{D}} \sum\limits_{i_2 \in \mathcal{D}} \left( \left| \sum\limits_{\tau \in \mathcal{H}} z_{i_{1}, \tau} - \sum\limits_{\tau \in \mathcal{H}}z_{i_{2}, \tau}\right|^{\frac{1}{2}} \right.$\\
        $- \left. \frac{1}{S} \sum\limits_{s=1}^{S} \left| \sum\limits_{\tau \in \mathcal{H}} \hat{z}_{i_1, \tau}^{(s)} - \sum\limits_{\tau \in \mathcal{H}} \hat{z}_{i_2, \tau}^{(s)} \right|^{\frac{1}{2}} \right)$} & Evaluated for summation over time periods, capturing spatial correlations. \\
        \hline
    \end{tabular}

    \label{tab:forecast_metrics}
\end{table*}

\subsection{Evaluation Metrics}
This section outlines various metrics employed to quantify the models for both deterministic and the probabilistic forecasting (i.e., scenario generation). Recall that, for a given target $i \in \mathcal{D}$ and each time step within the forecasting time horizon ($\tau \in \mathcal{H}$), the model produces output in the form of marginal CDFs, whereas $S$ forecasts (i.e., scenarios) are generated for the scenario generation using Algorithm 1. This essentially gives the prediction one more dimension, which indiced by superscript $(s)$.

The evaluation of the deterministic forecasts are performed at the individual and space-sum levels. Additionally, the probabilistic forecasts metrics are computed over time-sum levels to quantify the ability of the models in capturing temporal dependencies.  Let $\mathbf{1}$ be the column vector of all ones (with proper dimension), the comparison will be:

\begin{itemize}
    \item individual (\ind) level --- directly comparing the forecast vectors $\hat{\mathbf{z}}_{i,\tau}^{(s)}, s \in \{1, ..., S\}$ and the ground truth ${\mathbf{z}}_{i,\tau}$;
    \item space-sum (\mys) level --- comparison after summation over all locations, i.e., comparing $\mathbf{1}^{\rm T} \hat{\mathbf{z}}_{\cdot,\mathcal{H}}^{(s)}$ with $\mathbf{1}^{\rm T} {\mathbf{z}}_{\cdot,\mathcal{H}}$;
    \item time-sum (\myt) level --- comparison after summation over all time steps, i.e., comparing $\hat{\mathbf{z}}_{\cdot,\mathcal{H}}^{(s)}\mathbf{1}$ with ${\mathbf{z}}_{\cdot,\mathcal{H}}\mathbf{1}$.
\end{itemize}

\noindent
Table \ref{tab:forecast_metrics} outlines the metrics used to evaluate deterministic and probabilistic forecasts.
The precision of the deterministic forecasts is evaluated using normalized absolute error (NMAE) and root mean squared error (RMSE) over the individual forecasts and space-sum levels. These metrics are calculated for the load, wind power, and solar power generation over unseen test data across the different models discussed in Section \ref{sec: models}, with and without the inclusion of the weather covariates. 

To evaluate the efficacy of the models and the significance of integrating the copula method for realistic scenario generation, the results report the energy distance (ED) and variogram score (VS). ED is a proper scoring rule that measures the distance between the predicted CDF and the actual observations. It generalizes the continuous ranked probability score (CRPS) and is particularly useful for vector-valued forecasts in high-dimensional spaces \citep{gneiting2007strictly}. This paper considers ED due to its generalization capabilities for high-dimensional data. However, in scenario generation, both ED and CRPS fail to detect differences in correlation structures \citep{bjerregaard2021introduction}. Therefore, VS is introduced to quantify differences between the correlations of the scenarios and the ground truth. More details regarding the metrics are presented in \ref{Sec: Appendix: Evaluation metrics} in the Appendix.

\section{Results}
\label{sec: Real experiments}

This section presents the experimental results for the deterministic and probabilistic forecasting of the load, and the wind and solar power generation within MISO. Initially, the deterministic forecasting results for different models discussed in Section \ref{sec: models} are presented to compare the performance with and without the inclusion of weather data. Following this, the probabilistic forecasting results are discussed, highlighting the rational for incorporating the copula technique to restore the correlations. The experimental results are reported on distinct individual and aggregate levels.

\subsection{Deterministic Prediction}
\label{sec:sec:sec:deterministic forecast accuracy}
\begin{table*}[!htb]
\centering
\caption{Deterministic Forecast Evaluation. Performance of different models are evaluated on MISO dataset with and without inclusion of the weather covariates. The metrics are calculated at individual (i.e., $\ind$) and space-sum (i.e., $\mys$) levels.}
\label{tab:point-forecast}
\resizebox{\linewidth}{!}{%
\begin{tabular}{crrrrcrrrrcrrrr} 
\toprule
   & \multicolumn{4}{c}{load} 
   && \multicolumn{4}{c}{wind} 
   && \multicolumn{4}{c}{solar} \\
   \cmidrule(lr){2-5} \cmidrule(lr){7-10} \cmidrule(lr){12-15}
    & \multicolumn{2}{c}{$\mys$} & \multicolumn{2}{c}{$\ind$} 
   && \multicolumn{2}{c}{$\mys$} & \multicolumn{2}{c}{$\ind$}
   && \multicolumn{2}{c}{$\mys$} & \multicolumn{2}{c}{$\ind$} \\
    \cmidrule(lr){2-3} \cmidrule(lr){4-5}
    \cmidrule(lr){7-8} \cmidrule(lr){9-10}
    \cmidrule(lr){12-13} \cmidrule(lr){14-15}
    Model 
         & NMAE & RMSE & NMAE & RMSE 
        && NMAE & RMSE & NMAE & RMSE  
        && NMAE & RMSE & NMAE & RMSE  \\ 
    \midrule
% NREL & 2.3\% & 3708.0 & 3.3\% & 885.6 & 11.4\% & 2967.8& \textbf{15.3\%}& 16.9& \textbf{1.6\%} & \textbf{38.4} & \textbf{3.4\%} & \textbf{0.2} \\
 ARIMA 
         & 4.2\% &    6316.1 &   5.2\% &  1302.4 
        && 16.6\% & 4136.8 & 28.5\% & 25.4 
        && 9.0\% & 190.3 & 12.4\% & 0.7 \\
    Dlinear 
         & 2.8\% & 4211.2 & 4.2\% & 1003.1 
        &&  16.7\% &    4134.2 &  26.9\% &  24.6  
        &&  7.2\% &     181.0 &  10.0\% &   0.7  \\
    Nlinear 
         & 3.3\% & 4964.5 & 3.9\% & 994.9 
        && 16.2\% & 4030.6 & 26.0\% & 23.9 
        &&7.4\% &     182.5 &  10.2\% &   0.7  \\
    DeepAR 
         & 5.1\% & 7261.4 & 6.1\% & 1428.8 
        &&16.8\% &    4357.0 &  27.2\% &  25.3  
        &&   5.2\% &     168.9 &   8.2\% &   0.7 \\
    TFT 
         & 2.9\% & 4473.8 & 3.7\% & 954.5 
        && 16.7\% &    4335.3 &  27.0\% &  25.7 
        &&  5.5\% &     173.3 &   8.9\% &   0.7 \\ 
\midrule
    WI-Dlinear 
         & 2.7\% &    4149.3 &   3.8\% &   947.3  
        && 9.6\%&    2615.7 &  18.5\% &  18.5 
        && 4.8\% &  130.2 &   7.3\% &   0.6  \\
    WI-Nlinear 
         & 3.0\% &    4572.3 &   3.9\% &   984.6  
        && 9.7\% &    2633.1 &  18.4\% &  18.4 
        &&  5.0\% &     134.9 &   7.5\% &   0.6 \\
    WI-DeepAR 
         & 1.6\% &    2461.2 &   2.3\% &   596.4
        && 10.3\% &    2778.3 &  18.7\% &  19.4 
        &&  4.0\% &     139.7 &   6.3\% &   0.6  \\
    WI-TFT 
         & \textbf{1.1\%} & \textbf{1702.6} & \textbf{1.9\%} &\textbf{500.4} 
        && \textbf{7.9\%} & \textbf{2152.1} &  \textbf{16.1}\% &  \textbf{17.0}  
        && \textbf{3.1\%} &     \textbf{106.1} &   \textbf{5.8}\%  &   \textbf{0.5}  \\
\bottomrule
\end{tabular}%
}
\end{table*}

\begin{figure*}[!htb]
    \centering

    \includegraphics[width = 1\linewidth]{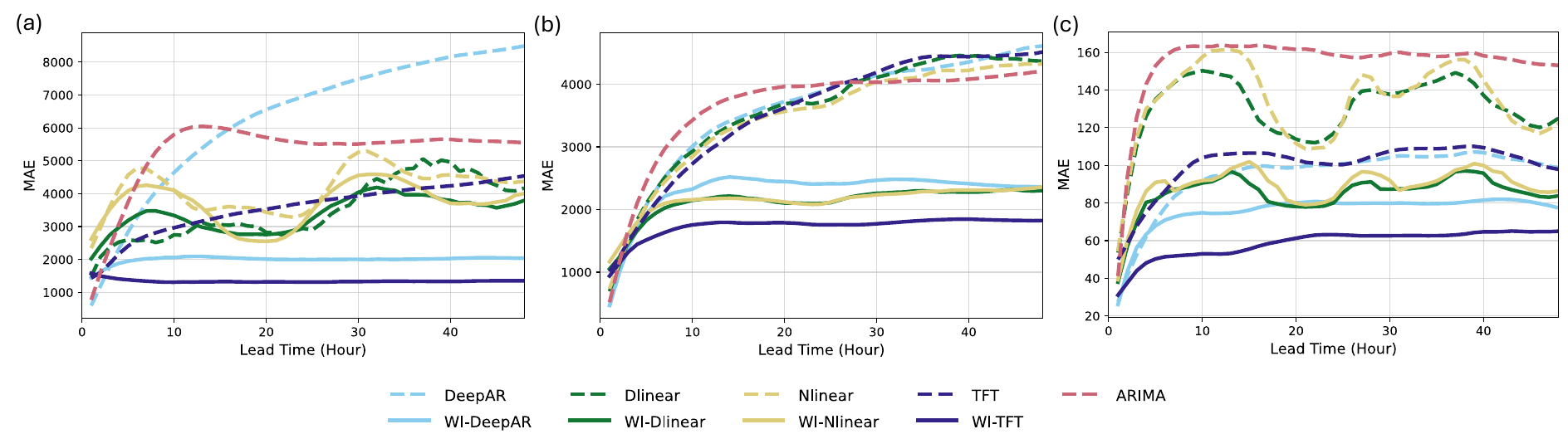}
    \caption{Total-level forecasting error (i.e., MAE) in MW versus lead time for the time series models for (a) load, (b) wind power generation, and (c) solar power generation.}
    \label{fig: err over time}
\end{figure*}
To evaluate the performance of the different models discussed in Section \ref{sec: models} for point-predictions and quantify the importance of weather covariates, deterministic forecasts are performed for each model. For each probabilistic forecasting method, and given the predicted marginal distributions of each space-time dimension, the expected values on each dimension are used as deterministic forecasts. For models that trained with quantile loss, the middle quantiles are used as deterministic forecasts, as they are optimized for L1 loss \citep{koenker2001quantile, wen2017multi}. The forecasting accuracy of each model is evaluated using $\text{NMAE}{\ind}$ and $\text{RMSE}{\ind}$ to measure the average forecast accuracy across all individual series. Additionally, $\text{NMAE}{\mys}$ and $\text{RMSE}{\mys}$ are used to assess the forecasts for total load, and solar and wind power generation at the aggregate space-sum level.

Table \ref{tab:point-forecast} presents the forecast errors for MISO system on the test dataset. The performance of models with and without weather forecast information is compared to examine the impact of weather covariates. The models that utilize weather data are denoted with the prefix \textit{WI} (i.e., weather informed) in Table \ref{tab:point-forecast}.

The comparison of deterministic forecasting methods shows how the DL-based models improve over the traditional ARIMA model. These improvements are primarily attributed to the transition from ARIMA’s univariate framework to a DL-based multivariate approach, which more accurately captures the complex dynamics within energy systems. The capacity of these models to effectively utilize cross-series information significantly enriches the forecasting process, offering a deeper understanding of energy patterns.
The integration of weather forecast information into these models further enhances their accuracy. It is observed that WI models consistently outperform those that do not include weather data, highlighting the importance of weather covariates in predicting the variability of RES. This is most pronounced in the TFT model, where incorporating weather covariates yields a significant reduction in prediction error by 49\% for load forecasting, 40\% for wind energy prediction, and 34\% for solar energy prediction on NMAE at an individual asset level. Similarly, adding covariantes improves RMSE at individual asset level by 30\% to 50\%. 

\begin{table*}[!htb]
\centering
\caption{Probabilistic Forecast Evaluation. Performance of the weather-informed (WI) models are evaluated on MISO dataset with and without inclusion of copula (i.e., mar. and cop., respectively) method for scenario generation. The metrics are calculated at space-sum (i.e., $\mys$), time-sum (i.e., $\myt$), and total levels. Lower values are better.}
\label{tab:prob_forecast}
\resizebox{\linewidth}{!}{%
\footnotesize
\begin{tabular}{clrrrrrrrr}
\toprule
    &
    & \multicolumn{2}{c}{ED$\mys$ } 
    & \multicolumn{2}{c}{VS$\mys$ }
    & \multicolumn{2}{c}{VS$\myt$ }
    & \multicolumn{2}{c}{ED } \\
    \cmidrule(lr){3-4} \cmidrule(lr){5-6} \cmidrule(lr){7-8} \cmidrule(lr){9-10}
Target & \multicolumn{1}{c}{Model}
    & \multicolumn{1}{c}{mar.} & \multicolumn{1}{c}{cop.} 
    & \multicolumn{1}{c}{mar.} & \multicolumn{1}{c}{cop.} 
    & \multicolumn{1}{c}{mar.} & \multicolumn{1}{c}{cop.} 
    & \multicolumn{1}{c}{mar.} & \multicolumn{1}{c}{cop.} \\
 \midrule
Load 
    & WI-DeepAR     & 25632.1 &        23882.4 &       655547.1 &              629964.7 &          27518.5 &                 \textbf{26209.7} &  15834.4 &    15791.6 \\
    & WI-Dlinear    & 46952.4 &        43919.8 &      1253299.7 &             1222333.8 &          65596.7 &                 63073.6 &   27159.3 &    27086.2 \\
    & WI-Nlinear    & 50969.3 &        46696.5 &      1452783.1 &             1406618.6 &          63242.8 &                 60534.6 &  27566.3 &    27528.2  \\
    & WI-TFT        & 17928.8 & \textbf{17553.7} &       343455.7 &              \textbf{341995.2} &          31011.6 &  30331.8 &  14299.4 &  \textbf{14277.0} \\ 
\midrule
Wind
    & WI-DeepAR &  29751.4 &        28845.1 &      1150740.6 &             1124880.2 &        7657926.0 &               7551511.4 &  5061.9 &     5036.3       \\
    & WI-Dlinear & 32004.6 &        26113.7 &      1329397.4 &             1152174.8 &        5038233.2 &               4877483.1 &  4571.7 &     4574.6  \\
    & WI-Nlinear & 32098.8 &        26033.3 &      1346814.8 &             1160315.4 &        4956035.6 &               4800996.2 &4522.9 &     4526.2 \\
    & WI-TFT & 25695.3 &        \textbf{24469.8} &      1062437.8 &             \textbf{1028368.5} &        4396148.6 &               \textbf{4268520.6 }&  \textbf{4452.2} &     4455.5 \\ 
\midrule
Solar
    & WI-DeepAR &  1509.4 &         1389.2 &        39688.5 &               39176.7 &         274696.7 &                275591.4 &  278.4 &      280.6 \\
    & WI-Dlinear &  1598.7 &         1275.8 &        51144.3 &               49122.6 &         235188.6 &                \textbf{229894.0} &  266.0 &      265.5\\
    & WI-Nlinear &  1657.9 &         1319.3 &        57429.9 &               54939.6 &         236787.0 &                230477.4 &  270.5 &      269.7 \\
    & WI-TFT &  1231.4 &         \textbf{1145.9} &        27296.2 &               \textbf{26915.7} &         259657.2 &                253595.5 & 263.9 &   \textbf{263.4}  \\ 
 \bottomrule
\end{tabular}%
}
\end{table*}

Moreover, the advantage of incorporating weather features becomes increasingly apparent for forecasts that extend further into the future, as depicted in Figure \ref{fig: err over time}. This trend is especially notable in wind energy forecasting: while models exhibit comparable accuracy for the initial 5-hour period, the performance of WI models remains robust beyond this window. In contrast, models lacking weather data experience a significant decline in performance over time. This pattern suggests that short-term predictions primarily rely on the current state of the system, whereas the accuracy of extended forecasts is increasingly contingent upon reliable weather information. A notable limitation of the linear models, such as DLinear and NLinear, is their inability to cope with the daily seasonality present in the data, which presents as a periodic pattern in the errors for load and solar power forecasting, as shown in Figure \ref{fig: err over time}. Conversely, the WI-TFT model maintains consistent performance across different targets and various lead times, highlighting its superiority in reliable forecasting for RES.

The NREL dataset also comes with predictions \citep{nrelperformdata}. Therefore, a comparison of prediction accuracy is made between the results of this paper and NREL. Every effort has been made to ensure a fair comparison. However, it should be noted that NREL forecasts have an 11-hour lead time and are updated once a day. In contrast, the forecasts in this paper are updated hourly without a lead time. This difference means that Table \ref{tab:point-forecast} contains 24 times more samples than the NREL forecasts. In the realm of load prediction, NREL forecasts exhibit a NMAE of 2.3\% and a RMSE of 3708.0 at the aggregate level, and a NMAE of 3.3\% alongside a RMSE of 885.6 at the individual level. The WI-TFT model for load prediction demonstrates a notable 52\% improvement in NMAE at the aggregate level and a 4.4\% improvement at the individual level over NREL predictions. When evaluating wind power predictions, the WI-TFT and NREL models show comparable performance at the individual level, with NREL registering a RMSE of 16.9. However, at the aggregate level, the WI-TFT model achieves a 44\% better NMAE than the 11.4\%from NREL. For solar power predictions, the WI-TFT model performs 70\% worse at the individual level and 94\% worse at the aggregate level in terms of NMAE compared to NREL. Despite this, both models exhibit high accuracy regarding the magnitude of errors, as evidenced in Table \ref{tab:point-forecast}, with the WI-TFT model averaging only a 0.5 MW error per generator based on RMSE at individual level.

\subsection{Scenario Generation}

\begin{figure*}[!htb]
    \centering
    \includegraphics[width = 0.5\linewidth]{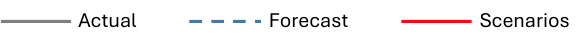}
    \vspace{-5pt}
    \vfill
    \begin{subfigure}[b]{0.4\linewidth}
        \centering
        \begin{subfigure}[b]{\linewidth}
            \centering
            \includegraphics[width=\linewidth]{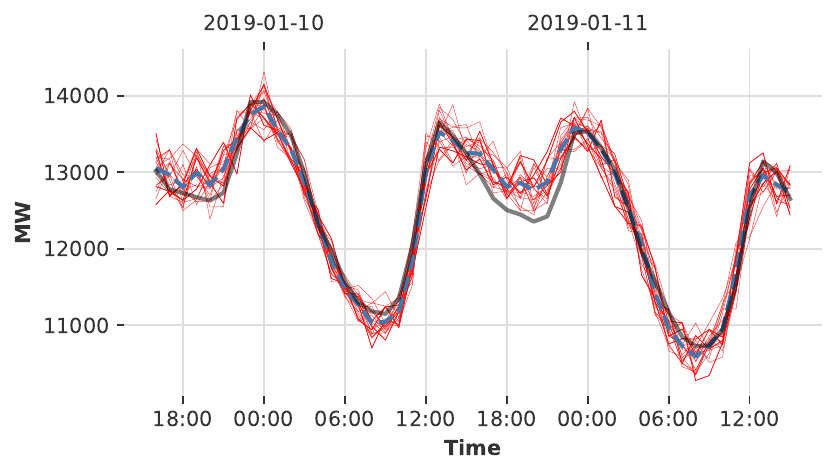}
            \caption{}
            \label{fig:margin-load-gen}
        \end{subfigure}
        \begin{subfigure}[b]{\linewidth}
            \centering
            \includegraphics[width=\linewidth]{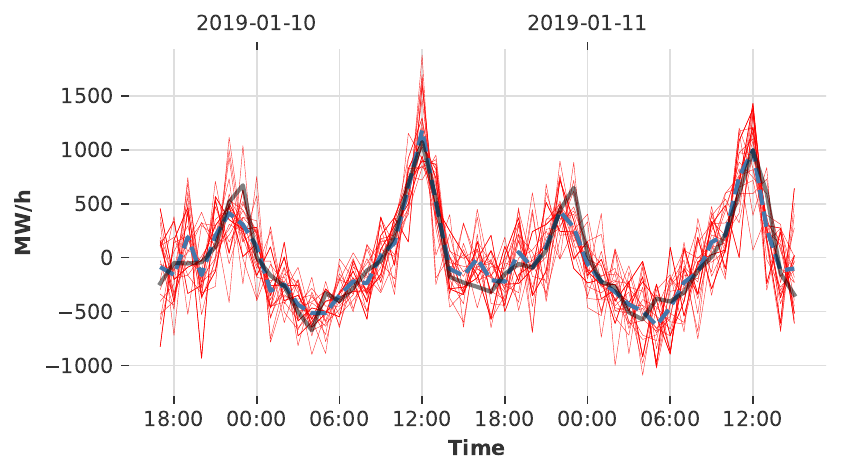}
            \caption{}
            \label{fig:load-mar-gen-ramps}
        \end{subfigure}
        \begin{subfigure}[b]{\linewidth}
            \centering
            \includegraphics[width=\linewidth]{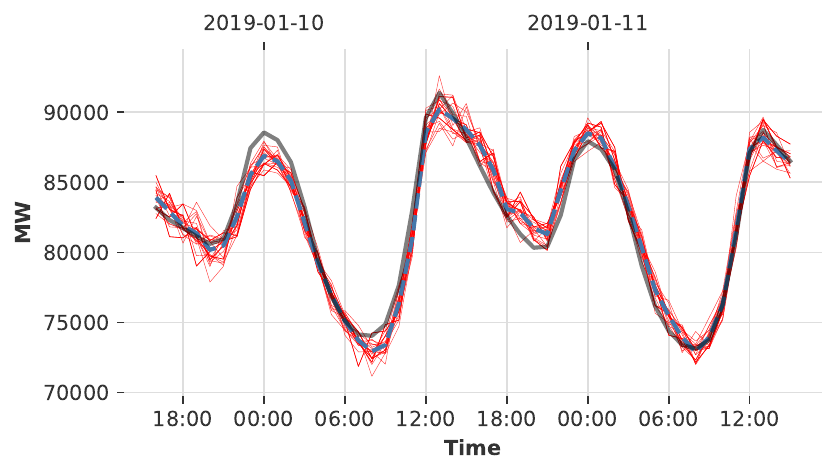}
            \caption{}
            \label{fig:margin-load-tot}
        \end{subfigure}
        \begin{subfigure}[b]{\linewidth}
            \centering
            \includegraphics[width=\linewidth]{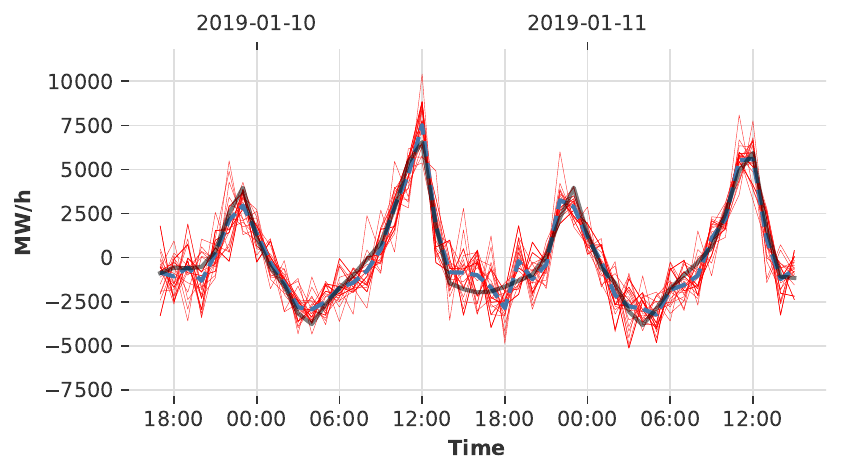}
            \caption{}
            \label{fig:load-mar-tot-ramps}
        \end{subfigure}
        \label{fig:marginal-scenarios}
    \end{subfigure}
    \begin{subfigure}[b]{0.4\linewidth}
        \centering
        \begin{subfigure}[b]{\linewidth}
            \centering
            \includegraphics[width=\linewidth]{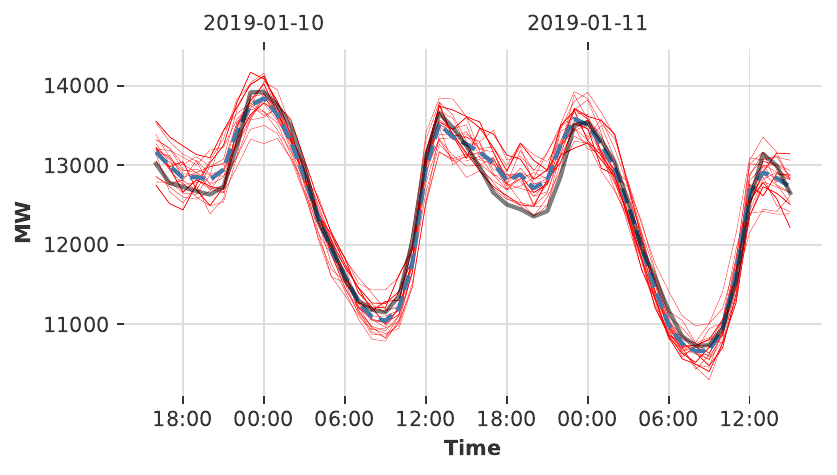}
            \caption{}
            \label{fig:copula-load-gen}
        \end{subfigure}
        \begin{subfigure}[b]{\linewidth}
            \centering
            \includegraphics[width=\linewidth]{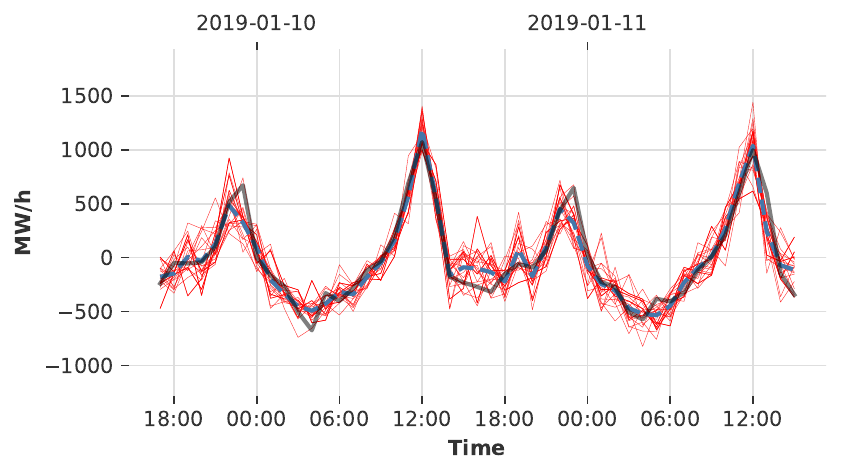}
            \caption{}
            \label{fig:load-copula-gen-ramps}
        \end{subfigure}
        \begin{subfigure}[b]{\linewidth}
            \centering
            \includegraphics[width=\linewidth]{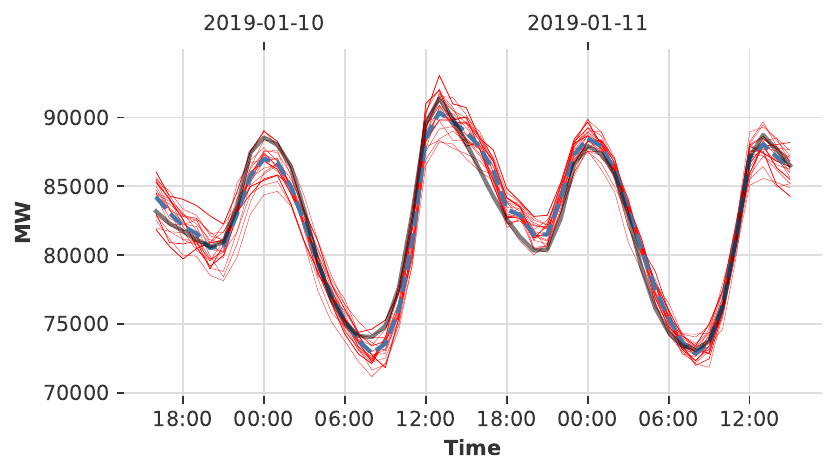}
            \caption{}
            \label{fig:copula-load-tot}
        \end{subfigure}
        \begin{subfigure}[b]{\linewidth}
            \centering
            \includegraphics[width=\linewidth]{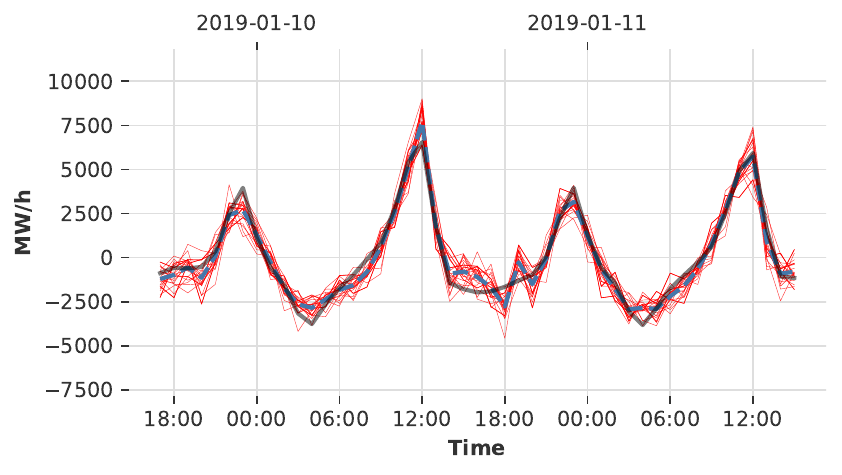}
            \caption{}
            \label{fig:load-copula-tot-ramps}
        \end{subfigure}
        \label{fig:copula-scenarios}
    \end{subfigure}
    \caption{Comparison of Marginal and Copula Scenarios. Generated scenarios for a load (LRZ1) using (a) marginals (i.e., no inclusion of copula) and (e) with the proposed copula method. (b) The ramps corresponding to the actuals and generated scenarios from marginals in (a). (f) The ramps for the actuals and generated scenarios using copula from (e). (c, d) The generated scenarios and the ramps for at total aggregated level using marginals and (g, h) using the proposed copula method.}
    \label{fig:copula-vs-marginal}
\end{figure*}
\begin{figure*}[!th]
\centering
    \includegraphics[width=1\linewidth]{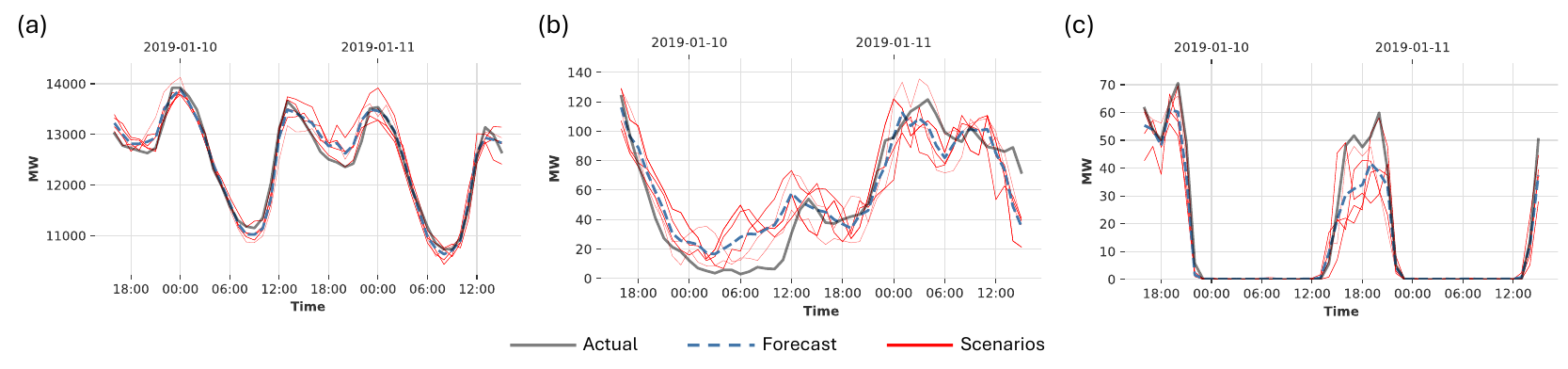}
\caption{WI-TFT + Copula for Scenario Generation. Examples of the forecast and 5 generated scenarios for (a) load, (b) wind, and (c) solar power generation. For simplicity, only 5 scenarios are plotted for each case.}
\label{fig:scenarios}
\end{figure*}

This section presents the results for the probabilistic forecasting of RES using the DL-based WI models discussed in the previous sections. Models that do not include weather covariates are omitted, as it was demonstrated that they consistently underperform compared to the WI models. Specifically, the performance of the generated scenarios for load, wind power, and solar power generation using different models are compared with and without integration of Gaussian copula. The evaluation metrics are computed for scenarios generated from both marginal distributions (mar.) and joint distributions represented by copula (cop.). For each model and individual target, 200 scenarios are generated (i.e., $S=200$) with and without the copula restoration. The performance of the models for generating scenarios are presented in Table \ref{tab:prob_forecast}.

The ED quantifies the distance between the distributions of the forecasted scenarios and the actual data. Lower ED values with the copula indicate improved alignment of generated scenarios with the true distribution. The ED for joint distribution and (\mys)-level are calculated and presented in Table \ref{tab:prob_forecast}. Notably, as the sample sizes remain constant and the dimension tends to infinity, ED can only identify the equality of means and the traces of covariance matrices \citep{chakraborty2021new}. This accounts for the absence of differences in ED in Table \ref{tab:prob_forecast}, since the copula does not affect the means and variances of the marginal distribution. 
Comparing the $\text{ED}_{\mys}$ with and without inclusion of the copula (i.e., cop. versus mar., respectively) shows that the copula model is better at capturing the joint distribution of the overall demand for the load and the overall wind and solar power generation with an improvement from $2\%$ to $20\%$ across different models.

The VS evaluates the correlation structure within the forecast scenarios. Due to the computational expense of VS for large dimensions, $\text{VS}_{\mys}$ and $\text{VS}_{\myt}$ are computed to capture the correlations across the spatial and temporal dimensions, respectively. The VS values are reduced by $0.4\%$ to $5\%$ upon integration of the copula, demonstrating its ability in improving the spatial and temporal correlations.

The WI-TFT model performs the best overall due to its superior ED and sufficiently low VS values. The lower ED indicates that the scenarios generated by WI-TFT are more accurate and closer to the true distribution, which is crucial for effective forecasting and decision making. It is important to note that VS does not capture the accuracy of the prediction and ED is more critical than VS in this context. Hence, ED and VS should be considered together for a comprehensive assessment of both the accuracy and reliability of the generated scenarios.

Figure \ref{fig:copula-vs-marginal} illustrates the comparative scenario analysis confirmed by the statistical data in Table \ref{tab:prob_forecast}. The left side of each plot displays various characteristics of scenarios generated from marginal distributions, while the right side depicts these characteristics after incorporating estimated correlations via the copula method. At the top of the figure, the plots \ref{fig:margin-load-gen} and \ref{fig:copula-load-gen} provide a 48-hour prediction for one load zone at the individual (\ind)-level. The coverage of the scenarios remains consistent before and after the application of the copula. However, the copula-generated scenarios exhibit smoother temporal correlations, aligning more closely with the ground truth. This smoother correlation is particularly evident in the Figures \ref{fig:load-mar-gen-ramps} and \ref{fig:load-copula-gen-ramps}, which shows the ramp rates of these scenarios. Here, the ramps of the copula models closely match the actual ramps, whereas those derived from marginal distributions show significant deviations.

Figures \ref{fig:copula-load-tot}, \ref{fig:margin-load-tot}, \ref{fig:load-mar-tot-ramps}, and \ref{fig:load-copula-tot-ramps} illustrate the same instance but at the space-sum (\mys)-level. Initially, the copula scenarios display wider coverage compared to the marginal scenarios, effectively capturing some essential spatial correlations. This enhanced coverage ensures that the copula scenarios more accurately encompass the actual total-level outputs. Similar to the 
(\ind)-level, the temporal correlations at the (\mys)-level are more accurately represented in the copula scenarios, as demonstrated by the improved ramp alignment in the final row of the figure.

The conclusions drawn from the load scenarios apply similarly to wind and solar. Figure \ref{fig:scenarios} shows the effectiveness of WI-TFT with copula in generating realistic scenarios for individual load zones, wind farms, and solar farms. The copula method improves temporal and spatial correlations, better aligning the generated scenarios with the actual data compared to marginal distributions alone. 

\section{Conclusion}
This paper proposed a new method based on combination of TFT and Gaussian copula for high-dimensional probabilistic forecasting in RES. Extensive experiments were conducted to compare different time series forecasting methods on a real-world forecasting problem within the MISO system. The results demonstrated superiority of the TFT model compared to other statistical and DL-based methods, and the significant impact of including weather information on the accuracy and effectiveness of the forecasts. The integration of weather data as covariates also improved the precision of the models over higher lead times. Additionally, the paper highlighted the efficacy of the Gaussian copula method in restoring spatio-temporal correlations, which is crucial for generating realistic scenarios in RES forecasting. This methodological advancement provides a robust framework for addressing the complexities associated with high-dimensional forecasting problems in RES.

Future work will focus on studying the effect of noise in covariates on the performance of the forecasts and scenarios. Investigating how inaccuracies in weather data influence the model's predictions is essential for further improving the robustness of the forecasting method. Additionally, uncertainty quantification will be considered using conformal prediction to provide confidence-aware predictions in high-dimensional RES forecasting problems. This will involve developing techniques to quantify and incorporate prediction uncertainty, thereby offering more reliable and actionable insights for system operators.

\printcredits
\FloatBarrier

\bibliographystyle{cas-model2-names}

\bibliography{references}

\appendix

\section{Evaluation Metrics}
\label{Sec: Appendix: Evaluation metrics}
In this section, the details of evaluation metrics are provided. 
\subsection{Deterministic Forecast}
In the case of point forecast, $S = 1$ and only one scenario/forecast is generated. For notational simplicity, the scenario index $s$ is dropped in this subsection.

\paragraph{Normalized Mean Absolute Error (NMAE)}
    For wind and solar power, the ground truth can sometimes become zero or close to zero. To avoid dividing by zeros, NMAE is considered. The first metric considers the forecast on each space-time dimension:
    \begin{equation}
        \text{NMAE}_{\ind} = \frac{1}{N_{\text{test}}DT} \sum\limits_{t=1}^{N_\text{test}} \sum\limits_{i \in \mathcal{D}, \tau \in \mathcal{H}} \left|\frac{\hat{z}_{i,\tau} - z_{i,\tau}}{\operatorname{max}\{\mathbf{z}_{i,\mathcal{T}_\text{train}}\}}\right|,
    \end{equation}
    expressed as a percentage. The $\text{NMAE}_{\ind}$ reflects the average performance on individual time series.
    
    The normalization value $z_{\operatorname{max}\{\mathbf{z}_{i,\mathcal{T}}\}}$ is the capacity of the generator if given. If the capacity is not given, it is the maximum value of the first year in the training dataset.
    
    The second metric considers summation over space and is defined as:
    \begin{equation}
        \text{NMAE}_{\mys} = \frac{1}{N_{\text{test}}T} \sum\limits_{t=1}^{N_\text{test}} \sum\limits_{\tau \in \mathcal{H}} \left|\frac{\sum\limits_{i=1}^{D} \hat{z}_{i,\tau} - \sum\limits_{i=1}^{D} z_{i,\tau}}{\sum\limits_{i=1}^{D} \operatorname{max}\{\mathbf{z}_{i,\mathcal{T}_\text{train}}\}}\right|.
    \end{equation}

\paragraph{Root Mean Square Error (RMSE)}
The RMSE is a commonly used metric to measure the accuracy of a model by calculating the square root of the average squared differences between the predicted and actual values. It provides a measure of the average magnitude of the error. The first metric considers the forecast on each space-time dimension:
\begin{equation}
    \text{RMSE}_{\ind} = \sqrt{\frac{1}{N_{\text{test}}DT} \sum\limits_{t=1}^{N_{\text{test}}}  \sum\limits_{i \in \mathcal{D}, \tau \in \mathcal{H}} \left(\hat{z}_{i,\tau} - z_{i,\tau}\right)^2},
\end{equation}
expressed in megawatts (MW). The $\text{RMSE}_{\ind}$ reflects the average performance on individual dimension.

The second metric considers summation over space and is defined as:
\begin{equation}
    \text{RMSE}_{\mys} = \sqrt{\frac{1}{N_{\text{test}}T} \sum\limits_{t=1}^{N_{\text{test}}} \sum\limits_{\tau \in \mathcal{H}} \left(\sum_{i=1}^{D} \hat{z}_{i,\tau} - \sum\limits_{i=1}^{D} z_{i,\tau}\right)^2},
\end{equation}
also expressed in MW. The $\text{RMSE}_{\mys}$ measures the model's accuracy in predicting the summation of wind power, solar power, or load over space.

\subsection{Probabilistic Forecast}

\paragraph{Energy Distance}
        The energy distance is a popular metric for evaluating the quality of probabilistic forecasts \citep{messner2020evaluation}.
        Given two probability distributions $\mathcal{F}$ and $\mathcal{G}$ with support in $\mathbb{R}^{D}$, the energy distance between $\mathcal{F} $ and $\mathcal{G}$ is defined as
        \begin{equation}
            \label{eq:M6.5:defs:energy_distance}
            \text{ED}(\mathcal{F}, \mathcal{G}) = 2 \mathbb{E} \left\| X - Y \right\| - \mathbb{E} \left\| X - X' \right\| - \mathbb{E} \left\| Y - Y' \right\|,
        \end{equation}
        where $\| \cdot \|$ is the Euclidean norm, and $X \sim \mathcal{F}, X' \sim \mathcal{F}, Y \sim \mathcal{G}, Y' \sim \mathcal{G}$ are independent random variables.
        In particular, when $\mathcal{F}$ and $\mathcal{G}$ are empirical probability distributions with realizations $\mathbf{x} = \{x_{1}, ..., x_{n} \}$ and $\mathbf{y} = \{y_{1}, ..., y_{m} \}$, an estimator of the energy distance using the U-statistics is given by:

        \begin{equation}
            \begin{split}
                \label{eq:defs:energy_distance:finite}
            \hat{\text{ED}}(\mathbf{x} , \mathbf{y}) = & \frac{2}{m n} \sum_{i=1}^{n} \sum_{j=1}^{m} \| x_{i} - y_{j} \|           \\
             & - \frac{1}{n^{2}} \sum_{i=1}^{n} \sum_{j=1}^{n} \| x_{i} - x_{j} \| - \frac{1}{m^{2}} \sum_{i=1}^{m} \sum_{j=1}^{m} \| y_{i} - y_{j} \|.
            \end{split}
        \end{equation}
        Note that, for $m=1$ case, the third term in the above summation is zero.

    Here, two energy distance metrics are considered:
    (i) At time step $t$, the $D$-by-$T$ matrix $z_{\cdot,\mathcal{H}}$ is vectorized by stacking the columns of the matrix on top of one another. The resulting $DT$-dimensional vector is denoted by $\operatorname{vec}(z_{\cdot,\mathcal{H}})$. There are $S$ generated forecasts, $\hat z_{\cdot,\mathcal{H}}^{(s)}, s = 1,\dots,S$ and the quality of those forecasts are measured by
    \begin{equation*}
    \begin{split}
        \hat{\text{ED}}&(\operatorname{vec}(z_{\cdot,\mathcal{H}}) , \{\operatorname{vec}(\hat z_{\cdot,\mathcal{H}})^{(s)}, s = 1,\dots,S\}) \\
        =& \frac{2}{S} \sum_{s=1}^S \left\|z_{\cdot,\mathcal{H}} - \hat z_{\cdot,\mathcal{H}}^{(s)}\right\|_F \\
        &- \frac{1}{S^2} \sum_{s_1, s_2 = 1}^S \left\| \hat z_{\cdot,\mathcal{H}}^{(s_1)} - \hat z_{\cdot,\mathcal{H}}^{(s_2)}\right\|_F,
    \end{split}
    \end{equation*}
    where $\|\cdot\|_F$ denotes the matrix $F$-norm and $\|\operatorname{vec}(\cdot)\| = \|\cdot\|_F$. In the following experiments, given $NS$ probabilistic forecasts produced, the average over $t = 1,\dots,N$ is taken as the first metric, which is also called the energy score
    \begin{equation}\label{eq:defs:energy_score}
    \begin{split}
        \text{ED}_{\ind} = \frac{1}{N_{\text{test}}} \sum\limits_{t \in \mathcal{N}_{\text{test}}} \hat{\text{ED}}(\operatorname{vec}(z_{\cdot,\mathcal{H}}) , \{\operatorname{vec}(\hat z_{\cdot,\mathcal{H}})^{(s)}, s = 1,\dots,S\}).
    \end{split}
    \end{equation}
    
 (ii) The energy distance computed on space-sum level is given by
\begin{equation*}
\begin{split}
    \hat{\text{ED}} & (\mathbf{1}^{\rm T} z_{\cdot,\mathcal{H}} ,  \{\mathbf{1}^{\rm T} \hat z_{\cdot,\mathcal{H}}^{(s)}, s = 1,\dots,S\}) \\
    =& \frac{2}{S} \sum_{s=1}^S \left\|\mathbf{1}^{\rm T} z_{\cdot,\mathcal{H}} - \mathbf{1}^{\rm T} \hat z_{\cdot,\mathcal{H}}^{(s)}\right\| \\
    &- \frac{1}{S^2} \sum_{s_1, s_2 = 1}^S \left\| \mathbf{1}^{\rm T} \hat z_{\cdot,\mathcal{H}}^{(s_1)} - \mathbf{1}^{\rm T} \hat z_{\cdot,\mathcal{H}}^{(s_2)}\right\|.
\end{split}
\end{equation*}        
The second metric, referred to as average space-sum energy score:
\begin{equation*}
\begin{split}
    \text{ED}_{\mys} = \frac{1}{N_{\text{test}}} \sum\limits_{t\in \mathcal{N}_{\text{test}}} \hat{\text{ED}}(\mathbf{1}^{\rm T} z_{\cdot,\mathcal{H}} , \{ \mathbf{1}^{\rm T} \hat z_{\cdot,\mathcal{H}}(s), s = 1,\dots,S\})
\end{split}
\end{equation*}

Recall that forecasts are produced for multiple time series and time periods.
Therefore, the energy distance in Eq. \eqref{eq:defs:energy_score} captures both the spatial and temporal quality of the forecasts.

    \paragraph{Variogram Score}

        The variogram score (VS) captures the correlations among multivariate quantities \citep{scheuerer2015variogram}.
        Given a probabilistic forecast at time $t$, the corresponding variogram score of order $p$ is defined as

        \begin{equation}
            \begin{split}
            \label{defs:variogram_score}
            \text{VS}_{p}(t)
            &= \sum_{i=1}^{D} \sum_{j=1}^{D} \sum_{\tau_{1} \in \mathcal{T}} \sum_{\tau_{2} \in \mathcal{T}} \big(
                \left| z_{i, \tau_{1}} - z_{j, \tau_{2}}\right|^{p} \\
            &    - \frac{1}{S} \sum_{s=1}^{S} \left| \hat{z}_{i, \tau_{1}}^{(s)} - \hat{z}_{j, \tau_{2}}^{(s)} \right|^{p}
            \big).
            \end{split}
        \end{equation}

        In all the experiments that follow, the variogram score is evaluated with $p=0.5$, following the recommendations in \citet{scheuerer2015variogram}.
        
        As can be seen from Eq \eqref{defs:variogram_score}, computing the variogram score for one time period has complexity $\mathcal{O}(D^{2}T^{2}S)$.
        This is computationally intractable for high dimensional setting, where $D \geq 300$, $T=48$, and $S=200$ for two days ahead solar power forecast.
        Therefore, variogram scores are reported for summation of load and renewable generation predictions over space. 
        This captures temporal correlations (between time periods), at the expense of spatial correlations (between time series).

        \begin{equation}
            \begin{split}
            \label{defs:variogram_score_1}
            \text{VS}_{p,\mys}(t)
            &=  \sum_{\tau_{1} \in \mathcal{T}} \sum_{\tau_{2} \in \mathcal{T}} \Bigg(
                \left| \sum_{i=1}^{D}  z_{i, \tau_{1}} - \sum_{i=1}^{D} z_{i, \tau_{2}}\right|^{p} \\
            &    - \frac{1}{S} \sum_{s=1}^{S} \left| \sum_{i=1}^{D} \hat{z}_{i, \tau_{1}}^{(s)} - \sum_{i=1}^{D} \hat{z}_{i, \tau_{2}}^{(s)} \right|^{p}
            \Bigg).
            \end{split}
        \end{equation}  
        
Averaging over the test set, the average space-sum variogram score is given as

\begin{equation}
    \text{VS}_{\mys} = \frac{1}{N_{\text{test}}} \sum_{t=1 }^{N_\text{test}} \text{VS}_{p,\mys}(t).
\end{equation}

        Another metric inter-changes the order of the summation and is defined as follows:

        \begin{equation}
            \begin{split}
            \label{defs:variogram_score_2}
            \text{VS}_{p,\myt}(t)
            &=   \sum_{i=1}^{D} \sum_{j=1}^{D} \big(
                \left| \sum_{\tau \in \mathcal{T}}  z_{i, \tau} - \sum_{\tau \in \mathcal{T}} z_{j, \tau}\right|^{p} \\
            &    - \frac{1}{S} \sum_{s=1}^{S} \left| \sum_{\tau \in \mathcal{T}} \hat{z}_{i, \tau}^{(s)} - \sum_{\tau \in \mathcal{T}} \hat{z}_{j, \tau}^{(s)} \right|^{p}
            \big).
            \end{split}
        \end{equation}

Averaging over the test set, the average time-sum variogram score is given as        
\begin{equation}
    \text{VS}_{\myt} = \frac{1}{N_{\text{test}}}\sum_{t=1 }^{N_\text{test}} \text{VS}_{p,\myt}(t).
\end{equation}

\subsection{Marginal Distributions}

When estimating marginal distributions, researchers typically employ either parametric or non-parametric approaches. This section introduces two prevalent methods: quantile regression and parameterized distribution estimation.

\paragraph{Non-Parametric Estimation: Quantile Regression} 
The quantile regression approach, as outlined in Wen et al.\citep{wen2017multi}, utilizes the concept of quantile loss in multistep multivariate settings. Consider $\mathcal{Q}$ to represent a predefined set of quantiles. For any given target quantity in space-time, denoted by $z_{i,\tau}$, our model generates a corresponding set of quantile estimates $\hat{z}_{i,\tau}^{(q)}$, where $q \in \mathcal{Q}$. The primary goal is to minimize the overall quantile loss function across all estimated quantiles, $\hat{z}_{i,\tau}^{(q)}$, formulated as follows:

    \begin{equation*}
        \min
            \sum_{t} \sum_{\tau \in \mathcal{H}} \sum_{q \in \mathcal{Q}} \sum_{i \in \mathcal{D}} \text{QL}_q\left(z_{i,\tau},\hat{z}_{i,\tau}^{(q)}\right),
    \end{equation*}
    where $t$ iterates over all forecast creation times. 

    The quantile loss (QL) is defined as 
    \begin{equation}
        \begin{split}                &\text{QL}_{q}\left(z_{i,\tau},\hat{z}_{i,\tau}^{(q)}\right) \\
            & \quad  =q\left(z_{i,\tau}-\hat{z}_{i,\tau}^{(q)}\right)^{+}+(1-q)\left(\hat{z}_{i,\tau}^{(q)}-z_{i,\tau}\right)^{+},
        \end{split}
        \label{eq: QL}
    \end{equation}
    where $(\cdot)^{+}=\max (0, \cdot)$.
    This paper considers the quantiles $q \in \{0.1,0.3,0.5,0.7,0.9\}$.

Once the quantiles $\hat{z}_{i,\tau}^{(q)}$ are available, the CDF function $\hat{F}_{i, \tau}(z_{i, \tau})$ can be estimated from fitting a smooth curve across the predicted quantiles. 

\paragraph{Parametric Estimation: Maximum Likelihood Estimation (MLE)}
The Beta MLE approach is utilized for estimating the parameters of a Beta distribution, typically denoted as $\alpha$ (alpha) and $\beta$ (beta). This method is particularly beneficial when the data is understood to follow a Beta distribution, often applicable to variables constrained between 0 and 1. For RES, it is common to scale the time series based on their capacity to capacity factors, therefore, beta distribution is chosen to demonstrate the parametric estimation. 

For a set of observations $z_{i,\tau}$ corresponding to each forecast time $\tau$ within the forecast horizon $\mathcal{H}$ and across all dimensions $i \in \mathcal{D}$, assuming these observations are drawn from a Beta distribution, the aim of Beta MLE is to find the parameter values that maximize the likelihood of the given sample.

The likelihood function for the Beta distribution, given the observations, is:

\[
\max_{\alpha_{i,\tau}, \beta_{i,\tau}}\prod_{t} \prod_{\tau \in \mathcal{H}} \prod_{i \in \mathcal{D}} \frac{z_{i,\tau}^{\alpha_{i,\tau} - 1} (1 - z_{i,\tau})^{\beta_{i,\tau} - 1}}{B(\alpha_{i,\tau}, \beta_{i,\tau})},
\]

where $t$ iterates all forecast creation times and $B(\alpha_{i,\tau}, \beta_{i,\tau})$ is the Beta function.

Optimization is typically performed on the log-likelihood,

\begin{equation}
    \begin{split}
          \max_{\alpha_{i,\tau}, \beta_{i,\tau}} \sum_{t} \sum_{\tau \in \mathcal{H}} \sum_{i \in \mathcal{D}} & [(\alpha_{i,\tau} - 1) \ln(z_{i,\tau}) + (\beta_{i,\tau} - 1) \ln(1 - z_{i,\tau})] \\ & - DH\ln(B(\alpha_{i,\tau}, \beta_{i,\tau})).  
    \end{split}
\end{equation}

The $\hat{\alpha}_{i,\tau}$ and $\hat{\beta}_{i, \tau}$ are found using gradient decent. Then the estimated marginal CDF functions can be written as $\hat{F}_{i,\tau} (z_{i,\tau}) = I_{z_{i,\tau}}( \alpha_{i,\tau} \beta_{i,\tau}) $ where $I_{z_{i,\tau}}( \alpha_{i,\tau} \beta_{i,\tau})$ is the regularized incomplete beta function.

\section{Baseline Models}

\subsection{Autoregressive Integrated Moving Average (ARIMA)}
    The ARIMA model is a popular and widely used statistical method for time series forecasting. It combines the concepts of autoregressive (AR), differencing (I), and moving average (MA) models to forecast future values based on past values. The AR component considers the linear relationship between an observation and a certain number of lagged observations. The I component involves differencing the series to make it stationary, removing trends or seasonality that can affect the model. However, the non-stationarity can be sometimes caused by a non-structured dynamic background \citep{wei2021inferring}; The method proposed therein can only handle time series retrospectively, and fails to forecast the dynamic background. The unstructured non-stationarity is out of the scope of the current work and is left for future discussion. The MA component accounts for the linear dependency between an observation and a residual error from a moving average model applied to lagged observations.

    The ARIMA model is a univariate model. The multivarite extention of ARIMA model is the Vector Autoregressive Integrated Moving Average (VARIMA) model. VARIMA models capture the dependencies between the lagged values of all the variables in the system. However, in the high-dimensional time series problem. The VARIMA model is unsuitable in practice because of the number of parameters in the linear model. Therefore, each time series is modeled independently using ARIMA model using the existing Auto-ARIMA package \citep{garza2022statsforecast}.  
    
\subsection{DeepAR}
\begin{figure}[ht]
    \centering
    \includegraphics[width=0.25\textwidth]{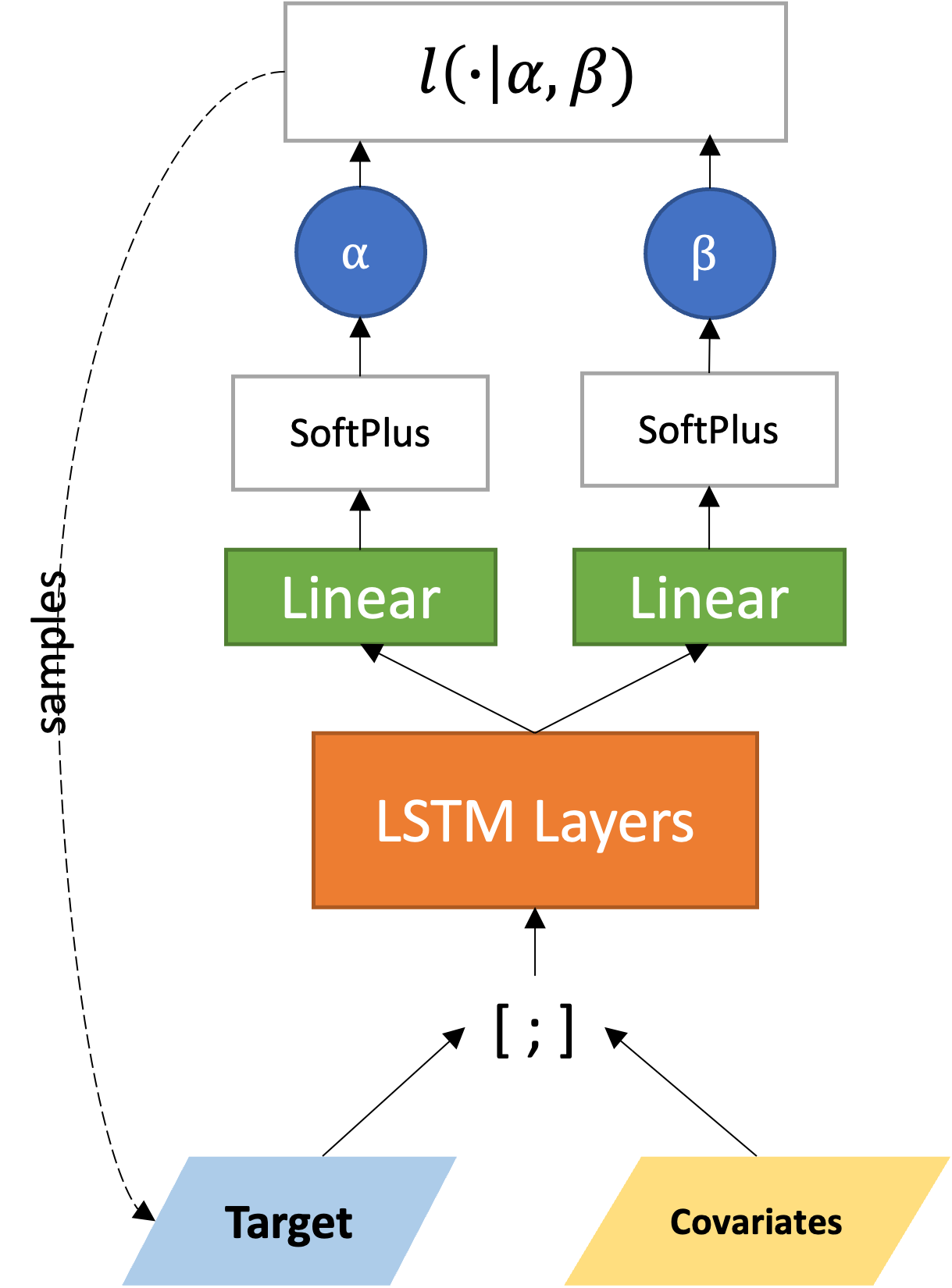}
    \caption{DeepAR}
    \label{fig:DeepAR}
\end{figure}
    DeepAR model is a deep learning-based approach that uses RNNs to capture temporal dependencies and can handle multiple related time series together. Similar to ARIMA model, DeepAR deals with multi-step ahead forecast by applying the model recursively, utilizing its own predictions as inputs for future predictions. As shown in Figure \ref{fig:DeepAR}, the model takes observations prior to time $t$ as input, and its predictions for the prediction horizon starting at time $t$ are used as inputs for the model at the next time step. The model's outputs are the parameters of a beta distribution, $\alpha$ and $\beta$, rather than a Gaussian distribution as used in the original paper. This is because the DeepAR model outputs capacity factors of RES that are bounded between 0 and 1, making the Gaussian distribution unsuitable.
        
    To estimate the model likelihood, the joint conditional distribution of ${Z}_{\cdot,\mathcal{T}}$ can be decomposed into a product of marginal conditional probabilistic distributions via Bayes rule as follows:
    \begin{equation}
         \prod_{t = W}^{N - H} \prod_{i \in \mathcal{D}, \tau \in \mathcal{H}} F^{(t)}_{i,\tau}\left(z_{i,\tau}^{(t)} \mid \mathbf{z}_{\cdot,\mathcal{W}}^{(t)},\mathbf{x}_{\cdot, \mathcal{W}}^{(t)}, \mathbf{x}_{\cdot, \mathcal{H}}^{(t)}\right).
    \end{equation}

    The marginal conditional distribution takes the following form
    \begin{equation*}
    \begin{split}
        & F^{(t)}_{i,\tau}\left(z_{i,\tau}^{(t)} \mid \mathbf{z}_{\cdot,\mathcal{W}}^{(t)},\mathbf{z}_{\cdot, \mathcal{W}}^{(t)}, \mathbf{x}_{\cdot, \mathcal{H}}^{(t)}\right)
        &=  \ell \left(z_{i,\tau}^{(t)} \mid \theta(\mathbf{h}_{i,t}^{(t)}; \Theta)\right),
    \end{split}
    \end{equation*}
    where the likelihood $\ell(\cdot)$ is parameterized by $\theta$, where $\theta$ depends on trainable parameter $\Theta$ and the output $\mathbf{h}_{i,t}^{(t)}$ of an autoregressive RNN:
    \begin{equation}\label{eq:hidden}
        \mathbf{h}_{i,t}^{(s)} = h \left(\mathbf{h}_{i,t-1}^{(t)}, \mathbf{z}_{\cdot,\mathcal{W}}^{(t)},\mathbf{z}_{\cdot, \mathcal{W}}^{(t)}, \mathbf{x}_{\cdot, \mathcal{H}}^{(t)} ; \Theta\right),
    \end{equation}
    and $h(\cdot)$ is a multi-layer RNN with LSTM cells. Here, ${\Theta}$ are typically associated with the neural network model architecture.
    This model is autoregressive in the sense that it exploits the observation at the last step $z_{i,t-1}$ and the previous output of the network $\mathbf{h}_{i,t-1}^{(t)}$ as an input at the next time step.

    In our work, because the amount of electricity generated by the wind farm is capped by the wind power capacity per wind farm, by normalizing with respect to the wind power capacity, the historical wind power time series can be represented within the range of $[0,1]$.
    Thus, the beta distribution for the target parametric distribution is more suitable.
    The likelihood of the beta distribution reads
    \begin{equation*}
        \ell(z; \alpha,\beta)=\frac{z^{\alpha-1}(1-z)^{\beta-1}}{B(\alpha,\beta)},  
    \end{equation*}
    where 
    \begin{equation}\label{eq:B-func}
        B(\alpha,\beta)=\frac{\Gamma(\alpha)\Gamma(\beta)}{\Gamma(\alpha+\beta)},
    \end{equation}
    and $\Gamma(\cdot)$ is the Gamma function.
    In this beta distribution example, 
        at time $t$, the model estimation can be done by maximizing the log-likelihood as follows:
        {
        \small
        \begin{equation*}
                        \max_{\bf{\Theta}}
             \frac{1}{(N-W-H) H D} \sum_{t=W}^{N - H} \sum_{i \in \mathcal{D},  \tau \in \mathcal{H}} \log \ell
                \left(
                    z^{(t)}_{i,\tau}; \alpha^{(t)}_{i, \tau},  \beta^{(t)}_{i, \tau}
                \right),
        \end{equation*}
        with $\theta(\mathbf{h}_{i,t}^{(t)}; \Theta) = \left(\alpha^{(t)}_{i, \tau},  \beta^{(t)}_{i, \tau}\right).$
        }

\subsection{NLinear and DLinear}
\begin{figure}[ht]
    \centering
    \includegraphics[width=0.4\textwidth]{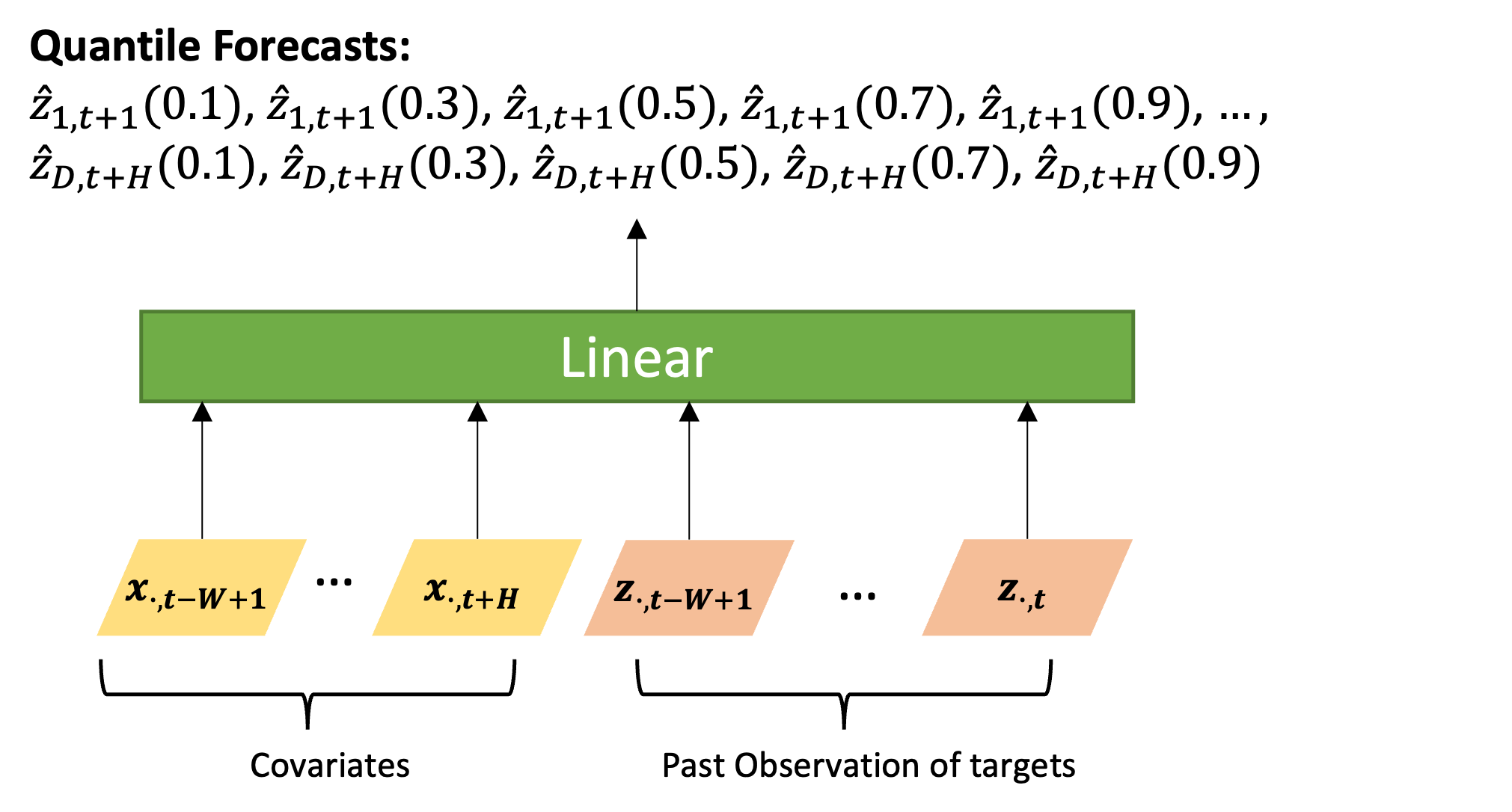}
    \caption{Linear Model}
    \label{fig:Nlinear}
\end{figure}

    The NLinear and Dlinear models are variants of the model described in \citet{zeng2022transformers}.
    The original paper does not consider the spatial correlations and covariates. And the weights are shared for every time series.
    To account for covariates (weather forecast), every future dimension in space and time is regressed on features of all past observations for all time series as well as covariates that extends to future.
    
    Figure \ref{fig:Nlinear} provides a visual representation of the models inputs and outputs for a vanilla linear (one-layer) model.
    Dlinear and Nlinear correspond to two preprocessing methods build upon the vanilla linear model. Dlinear uses decomposition scheme, which decomposes the input into trend and seasonal component.
    NLinear normalizes each input by subtracting the last value in the sequence, and then adds it back after passing through the linear layer to address distribution shift. Initially, the models were only used for point predictions. However, in this paper, it is adapted to a probabilistic setting by using quantile regression as the objective function, as described in the TFT \citep{LIM20211748} model.

\subsection{Temporal Fusion Transformer (TFT)}
\begin{figure*}[ht]
    \centering
    \includegraphics[width = 0.8\textwidth]{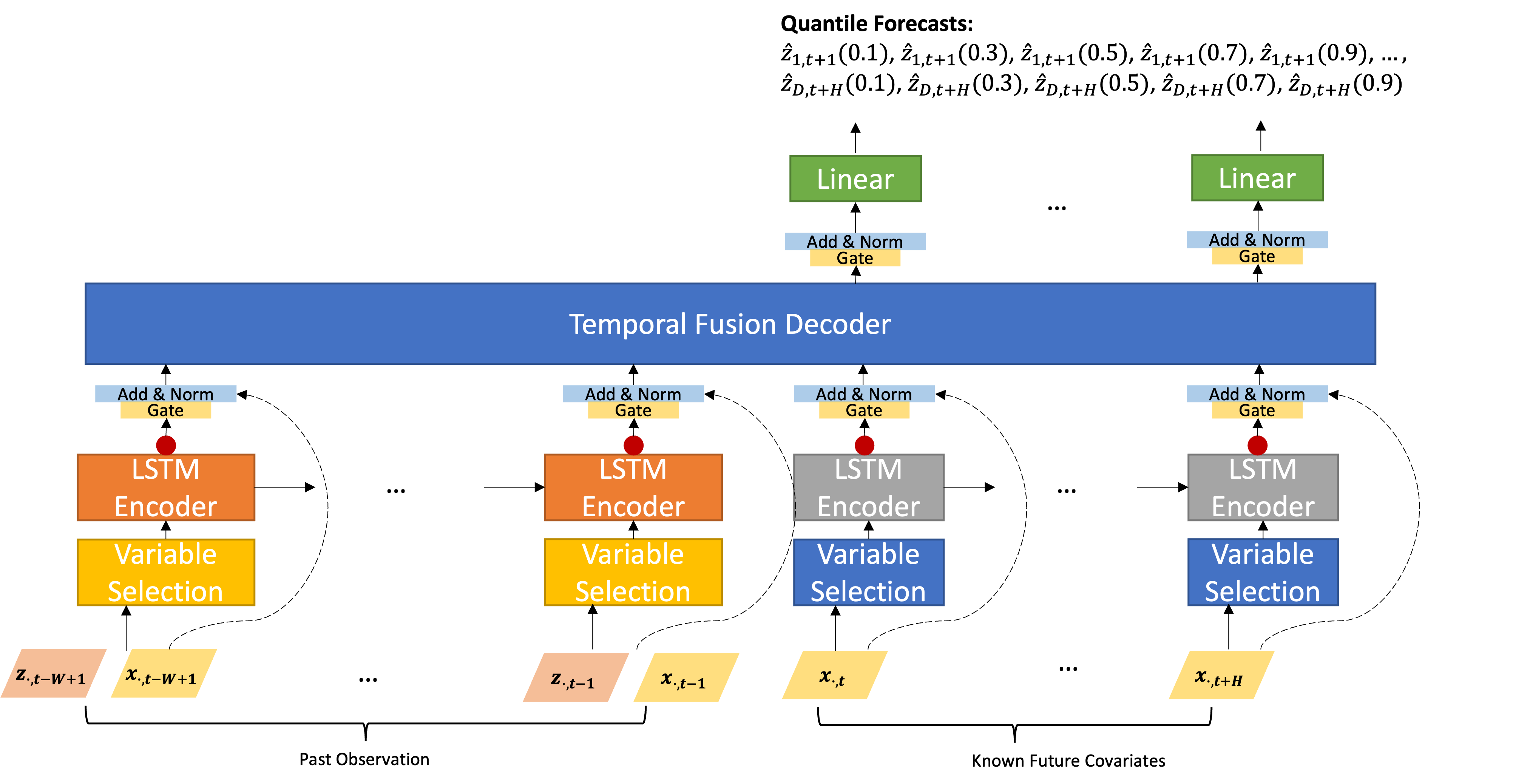}
    \caption{Temporal Fusion Transformer (TFT)}
    \label{fig:TFT}
\end{figure*}
    TFT is a deep learning model designed for multi-horizon forecasting tasks with mixed-type inputs, including static covariates, known future inputs, and past exogenous time series.
    TFT uses a combination of recurrent layers and self-attention layers to learn temporal relationships at different scales and provide interpretable insights into temporal dynamics.
    The model utilizes specialized components to select relevant features and gating layers to suppress unnecessary components, resulting in high performance in a wide range of scenarios.
    Figure \ref{fig:TFT} presents an illustration of TFT \citep{LIM20211748} architecture. 

    TFT models output prediction intervals via a quantile regression approach.
    For a chosen set of quantiles $\mathcal{Q}$ with $|\mathcal{Q}| = Q$, the TFT model outputs forecast quantiles $\hat{z}_{i,\tau}^{(t)}(q), \  q \in \mathcal{Q}$.
    The objective is to minimize the quantile loss function with respect to all quantile outputs $\hat{z}_{i,\tau}^{(t)}(q)$'s, i.e.,
    {\small
    \begin{equation*}
        \min
            \frac{1}{(N-W-H) H D} \sum_{t=W}^{N - H} \sum_{i \in \mathcal{D},  \tau \in \mathcal{H}} \sum_{q \in \mathcal{Q}} \text{QL}\left(z_{i,\tau}^{(t)},\hat{z}_{i,\tau}^{(t)}(q)\right),
    \end{equation*}
    }
    where
    \begin{equation*}
        \begin{split}                &\text{QL}\left(z_{i,\tau}^{(s)},\hat{z}_{i,\tau}^{(s)}(q)\right) \\
            & \quad  =q\left(z_{i,\tau}^{(s)}-\hat{z}_{i,\tau}^{(s)}(q)\right)^{+}+(1-q)\left(\hat{z}_{i,\tau}^{(s)}(q)-z_{i,\tau}^{(s)}\right)^{+},
        \end{split}
    \end{equation*}
    
    and $(\cdot)^{+}=\max (0, \cdot)$.
    All experiments use values of $q \in \{0.1,0.3,0.5,0.7,0.9\}$.
    In the inference phase, scenarios are sampled by first generating uniformly-distributed noised in $[0, 1]$, then by applying the inverse CDF defined by the quantiles produced by the TFT model.

\section{Training Process}
\label{appendix:sec:trainingprocess}
The models were trained using the Adam optimizer, and a learning rate scheduler was employed to adaptively adjust the learning rate during training. Specifically, we utilized the \texttt{ReduceLROnPlateau} scheduler with a reduction factor of 0.1, a threshold of 0.0001, a patience of 5 epochs, and a minimum learning rate of \(\text{lr} \times 10^{-4}\). Early stopping criteria were also implemented based on the validation loss, with a patience of 10 epochs and a minimum delta of 0.0001 to prevent overfitting. The detailed configurations of each model are provided in Table~\ref{tab:model_configurations}. All experiments were conducted on a single node equipped with one NVIDIA A100 GPU using the PACE cluster infrastructure~\cite{PACE}.

\begin{table}[h!]
\centering
\resizebox{\columnwidth}{!}{
\begin{tabular}{ccrrr}
\hline
\textbf{Model} & \textbf{Parameter} & \textbf{Load} & \textbf{Solar} & \textbf{Wind} \\
\hline
\multirow{5}{*}{WI-DeepAR} & in\_len & 22 & 24 & 23 \\
 & hidden\_dim & 256 & 191 & 234 \\
 & n\_rnn\_layers & 1 & 1 & 1 \\
 & dropout & 0.2635 & 0.3275 & 0.1677 \\
 & lr & 0.000825 & 0.003539 & 0.003172 \\
\hline
\multirow{3}{*}{WI-DLinear} & in\_len & 12 & 8 & 4 \\
 & kernel\_size & 10 & 7 & 2 \\
 & lr & 0.000097 & 0.000063 & 0.000177 \\
\hline
\multirow{2}{*}{WI-NLinear} & in\_len & 10 & 6 & 1 \\
 & lr & 0.000126 & 0.000084 & 0.000218 \\
\hline
\multirow{7}{*}{WI-TFT} & in\_len & 22 & 14 & 11 \\
 & hidden\_size & 45 & 101 & 118 \\
 & num\_attention\_heads & 3 & 4 & 2 \\
 & hidden\_continuous\_size & 87 & 39 & 40 \\
 & lstm\_layers & 2 & 2 & 2 \\
 & dropout & 0.23 & 0.21 & 0.36 \\
 & lr & 0.0073 & 0.0022 & 0.0032 \\
\hline
\end{tabular}
}
\caption{Model configurations for load, solar, and wind}
\label{tab:model_configurations}
\end{table}

\end{document}